\documentclass[sigconf, nonacm]{acmart}

\AtBeginDocument{%
  }

\usepackage{subcaption}
\usepackage{xspace}
\usepackage{xcolor}
\usepackage{adjustbox}

\usepackage{amssymb}
\usepackage{pifont}
\usepackage{cuted}
\newcommand{\xmark}{\ding{55}}%
\usepackage{hyperref}
\usepackage{paralist} 

\usepackage[utf8]{inputenc}
\usepackage{pgf-pie}
\usepackage{soul}
\usepackage{svg}
\usepackage{makecell}

\newcommand{\dataset}{\texttt{CartoMapQA}\xspace}

\newcommand{\tasks}{objectives\xspace}
\newcommand{\subtask}{task\xspace}
\newcommand{\subtasks}{tasks\xspace}
\newcommand{\question}{request\xspace}
\newcommand{\questions}{requests\xspace}
\newcommand{\map}{cartographic map\xspace}

\newcommand{\maps}{cartographic maps\xspace}

\newcommand{\oAI}{OpenAI}
\newcommand{\oone}{o1}
\newcommand{\othree}{o3}
\newcommand{\gemini}{\acl{gemini}}
\newcommand{\claude}{\acl{claude}}
\newcommand{\gptV}{GPT-4(V)}
\newcommand{\gptO}{GPT-4o}

\newcommand{\OAI}{\emph{\oAI}}
\newcommand{\Oone}{\emph{\oone}}
\newcommand{\Othree}{\emph{\othree}}
\newcommand{\GeminiP}{\emph{\gemini}}
\newcommand{\Claude}{\emph{\claude}}

\newcommand{\GptO}{\emph{\gptO}}

\newcommand{\llama}{\acl{llama} 3.2\xspace}
\newcommand{\llamaS}{\acl{llama} 4 Scout\xspace}
\newcommand{\intern}{\acl{intern} 2.5\xspace}
\newcommand{\llava}{\acs{llava}\xspace}
\newcommand{\qwen}{\acs{qwen} 2.5\xspace}

\newcommand{\Llama}{\emph{\llama}\xspace}
\newcommand{\LlamaS}{\emph{\llamaS}\xspace}
\newcommand{\Intern}{\emph{\intern}\xspace}
\newcommand{\Llava}{\emph{\llava}\xspace}
\newcommand{\Qwen}{\emph{\qwen}\xspace}

\usepackage{marvosym}  
\usepackage{graphicx}
\usepackage{caption}
\usepackage{stfloats} 
\usepackage{multirow}
\usepackage{tabularx}

\usepackage[printonlyused]{acronym} 
\usepackage{hyperref} 

\acrodef{llt}[LLT]{low-level task}
\acrodef{hlt}[HLT]{high-level task}
\acrodef{agi}[AGI]{Artificial General Intelligence}
\acrodef{cv}[CV]{Computer Vision}
\acrodef{cm}[CM]{Cartographic Map}
\acrodef{eu}[EU]{End-user}
\acrodef{mapfeat}[MF]{Map Feature}
\acrodef{mfsem}[MFS]{Map Feature Semantics}
\acrodef{stmf}[STMF]{Single-Type Map Feature}
\acrodef{mtmf}[MTMF]{Multiple-Type Map Feature}
\acrodef{rlest}[RLE]{Route Length Estimation}
\acrodef{srnav}[SRN]{Shortest Route Navigation}
\acrodef{mmloc}[MML]{Map Marker Localization}
\acrodef{llm}[LLM]{Large Language Model}
\acrodef{mllm}[MLLM]{Multi-modal Large Language Model}
\acrodef{vlm}[LVLM]{Large Visual-Language Model}
\acrodef{poi}[POI]{Point of Interest}
\acrodef{ocr}[OCR]{Optical Character Recognition}
\acrodef{cot}[CoT]{chain-of-thought}
\acrodef{vqa}[VQA]{Visual Question Answering}
\acrodef{suppm}[SM]{Supplementary Material}
\acrodef{rmse}[RMSE]{Root Mean Squared Error}
\acrodef{mape}[MAPE]{Mean Absolute Percentage Error}
\acrodef{gis}[GIS]{Geographic Information Systems}

\acrodef{gemini}{Gemini 2.5 Pro}
\acrodef{claude}{Claude 3.7 Sonnet}
\acrodef{llama}{LLama}
\acrodef{intern}{InternVL}
\acrodef{qwen}[Qwen-VL]{Qwen-VisionLanguage}
\acrodef{llava}[LLaVa-OV]{LLaVa-OneVision}

\copyrightyear{2025}
\acmYear{2025}
\setcopyright{acmlicensed}
\acmConference[SIGSPATIAL '25]{The 33rd ACM International Conference on Advances in Geographic Information Systems}{November 3--6, 2025}{Minneapolis, MN, USA} 
\acmBooktitle{The 33rd ACM International Conference on Advances in Geographic Information Systems (SIGSPATIAL '25), November 3--6, 2025, Minneapolis, MN, USA} 
\acmDOI{10.1145/3748636.3762755} 
\acmISBN{979-8-4007-2086-4/2025/11}

\begin{document}

\title[CartoMapQA]{CartoMapQA: A Fundamental Benchmark Dataset Evaluating Vision-Language Models on Cartographic Map Understanding}

\author{Huy Quang Ung}
\email{xhu-ung@kddi.com}
\orcid{0000-0001-9238-8601}
\affiliation{%
  \institution{KDDI Research, Inc.}
  \city{Fujimino}
  \country{Japan}
}
\author{Guillaume Habault}
\orcid{0000-0002-3364-5863}
\email{xgu-habault@kddi.com}
\affiliation{%
  \institution{KDDI Research, Inc.}
  \city{Fujimino}
  \country{Japan}
}
\author{Yasutaka Nishimura}
\orcid{0000-0003-4487-6285}
\email{yu-nishimura@kddi.com}
\affiliation{%
  \institution{KDDI Research, Inc.}
  \city{Fujimino}
  \country{Japan}
}

\author{Hao Niu}
\email{ha-niu@kddi.com}
\orcid{0000-0002-5623-9470}
\affiliation{%
  \institution{KDDI Research, Inc.}
  \city{Fujimino}
  \country{Japan}
}

\author{Roberto Legaspi}
\email{ro-legaspi@kddi-research.jp}
\orcid{0000-0001-8909-635X}
\affiliation{%
  \institution{KDDI Research, Inc.}
  \city{Fujimino}
  \country{Japan}
}

\author{Tomoki Oya}
\email{to-ooya@kddi.com}
\orcid{0009-0005-4927-076X}
\affiliation{%
  \institution{KDDI Research, Inc.}
  \city{Fujimino}
  \country{Japan}
}

\author{Ryoichi Kojima}
\email{ry-kojima@kddi.com}
\orcid{0009-0009-3128-7781}
\affiliation{%
  \institution{KDDI Research, Inc.}
  \city{Fujimino}
  \country{Japan}
}

\author{Masato Taya}
\email{ma-taya@kddi.com}
\orcid{0009-0006-4911-4289}
\affiliation{%
  \institution{KDDI Research, Inc.}
  \city{Fujimino}
  \country{Japan}
}

\author{Chihiro Ono}
\email{ci-ono@kddi.com}
\orcid{0000-0002-6410-1359}
\affiliation{%
  \institution{KDDI Research, Inc.}
  \city{Fujimino}
  \country{Japan}
}

\author{Atsunori Minamikawa}
\email{at-minamikawa@kddi.com}
\orcid{0009-0009-8856-7813}
\affiliation{%
  \institution{KDDI Research, Inc.}
  \city{Fujimino}
  \country{Japan}
}

\author{Yan Liu}
\email{yanliu.cs@usc.edu}
\orcid{0000-0002-7055-9518}
\affiliation{%
  \institution{University of Southern California}
  \city{Los Angeles}
  \country{USA}
}

\renewcommand{\shortauthors}{Ung et al.}


\begin{abstract}
The rise of Visual-Language Models (LVLMs) has unlocked new possibilities for seamlessly integrating visual and textual information. 
However, their ability to interpret cartographic maps remains largely unexplored.
In this paper, we introduce \texttt{CartoMapQA}, a benchmark specifically designed to evaluate LVLMs' understanding of cartographic maps through question-answering tasks.
The dataset includes over $2000$ samples, each composed of a cartographic map, a question (with open-ended or multiple-choice answers), and a ground-truth answer. 
These tasks span key low-, mid- and high-level map interpretation skills, including symbol recognition, embedded information extraction, scale interpretation, and route-based reasoning.
Our evaluation of both open-source and proprietary LVLMs reveals persistent challenges: models frequently struggle with map-specific semantics, exhibit limited geospatial reasoning, and are prone to Optical Character Recognition (OCR)-related errors. 
By isolating these weaknesses, \texttt{CartoMapQA} offers a valuable tool for guiding future improvements in LVLM architectures. 
Ultimately, it supports the development of models better equipped for real-world applications that depend on robust and reliable map understanding, such as navigation, geographic search, and urban planning.
Our source code and data are openly available to the research community at:\\\textit{\url{https://github.com/ungquanghuy-kddi/CartoMapQA.git}}
\end{abstract}



\begin{teaserfigure}
  \centering
  \includegraphics[width=0.95\textwidth]{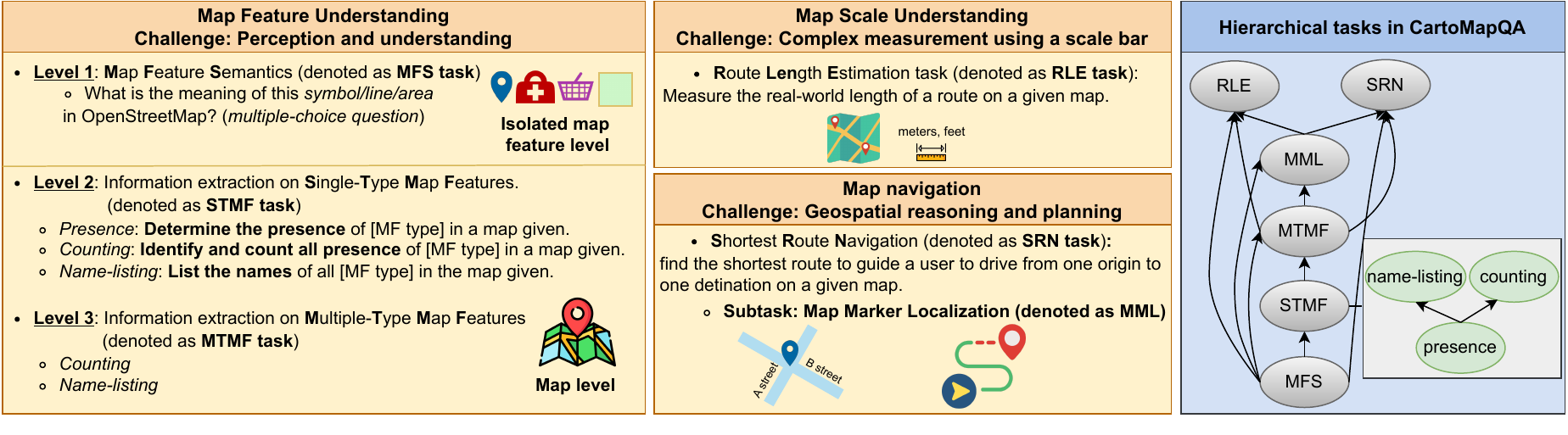}
    \caption{Overview of \texttt{CartoMapQA}, a benchmark  designed to evaluate Large Visual-Language Models on cartographic map understanding, including Map Feature recognition, map scale interpretation, and turn-by-turn navigation. It consists of six hierarchically structured tasks, where lower-level tasks serve as foundation for more complex, higher-level ones.
    }
    \Description{Overview image of the benchmark for evaluating VLMs on map understanding.}
  \label{fig:teaser}
\end{teaserfigure}

\maketitle

\section{Introduction}

Recent advancements in \acp{llm} and \acfp{vlm} have demonstrated remarkable generalization and reasoning capabilities across a variety of domains, including programming~\cite{hongmetagpt}, mathematical reasoning~\cite{ahn2024large}, \ac{vqa}~\cite{liu2023visual}, and common sense reasoning~\cite{suzgun2023challenging}. 
More recently, there has been growing interest in leveraging \acp{llm} and \acp{vlm} for geospatial applications~\cite{bhandari2023large, mooney2023towards, roberts2023gpt4geo, gurneelanguage, deng2024k2}, with work spanning both model development~\cite{yan2024urbanclip,manvigeollm, bhandari2023large} and the construction of specialized benchmarks~\cite{lobry2020rsvqa, feng2024citybench,yang2024v,dihan2024mapeval}. 
These efforts target key tasks such as object recognition~\cite{feng2024citybench}, counting~\cite{zhang2024good}, and geospatial reasoning, including social indicator estimation~\cite{manvigeollm, yan2024urbanclip} and location-based information retrieval~\cite{dihan2024mapeval}.
Despite their promise, existing approaches often rely on text descriptions or satellite imagery~\cite{yan2024urbanclip,kuckreja2024geochat}, each with critical limitations. 
Textual data is constrained by token limits and typically lacks the spatial richness needed for detailed geospatial reasoning. 
Satellite imagery, while rich in visual detail, frequently includes irrelevant elements---such as foliage or architectural variation---that may distract models from the task at hand.

\Acp{cm} present a compelling alternative. 
Like satellite imagery, they offer spatially grounded visual input, but in a more structured and abstracted form. 
Using standardized symbols, color-coded regions, and textual labels, \acp{cm} provide a simplified yet information-rich view of geographic data---making them especially well suited for \acp{vlm} to reason about spatial relationships, landmarks, distances, and navigation.
Recently, Dihan et al.~\cite{dihan2024mapeval} has evaluated \acp{mllm} using \acp{cm}, however their benchmark focused mainly on user-level complex tasks, limiting potential insight into foundational limitations of \acp{vlm}. 
Moreover, their experiments with visual inputs---based on proprietary \emph{Google Maps} imagery---showed lower performance compared to selective textual context or API inputs, underscoring the difficulty of map-based reasoning.

In this paper, we present \dataset, a benchmark dataset specifically designed to evaluate \acp{vlm}' ability to interpret and reason over \maps through a structured set of question-answering tasks. 
Figure~\ref{fig:teaser}, \dataset defines three \tasks:
\begin{inparaenum}
    \item \textbf{Map Feature Understanding}: Identify (\ac{mfsem} \subtask), detect, count, and associate \ac{mapfeat} with corresponding labels (\ac{stmf} and \ac{mtmf} \subtasks).
    \item \textbf{Map Scale Understanding}: Locate the scale notation, identify pre-defined routes, measure route length and compute real-world distances between map markers (\ac{rlest} \subtask).
    \item \textbf{Map Navigation}: Comprehend direction, determine the shortest drivable route between map markers and describe turn-by-turn navigation (\ac{srnav} \subtask), with supporting evaluation on marker localization accuracy (\ac{mmloc} \subtask).
\end{inparaenum}
The dataset consists of $2251$ questions across $853$ maps sourced from \emph{OpenStreetMap}~\cite{OpenStreetMap} across three English-speaking countries, ensuring that language comprehension does not influence the evaluation. 
We adopt \emph{OpenStreetMap}’s definition of \ac{mapfeat}---any element represented by a symbol, line, or colored area, possibly with a textual label such as \acp{poi}, roads, or land-use areas.

To assess model capabilities with \maps, we evaluate various \acp{vlm} in a zero-shot setting. 
Our findings include:
While detailed results are presented in the subsequent sections, our findings include:
\begin{itemize}
    \item While proprietary models like \GeminiP{}~\cite{gemini2.5}, \OAI{} \Oone~\cite{jaech2024openai}, and \Othree~\cite{o3} achieved top scores, across all \subtasks, their performance is still limited.
    \item Open-source models generally underperform across most \subtasks, with significant performance gaps in higher-level reasoning \subtasks.
    \item In both \ac{stmf} and \ac{mtmf} \subtasks, models often correctly associate \acp{mapfeat} with their textual labels. Error analysis reveals minimal hallucination, with most failures stemming from \ac{ocr} error or the incorrect association of a \ac{mapfeat} with a semantically related type (e.g., restaurant vs. fast-food).
    \item In \ac{rlest}, a review of \Oone's answers on some samples shows promising results but with irregular behaviors.
    \item \ac{srnav} remains particularly challenging; the best-performing \ac{vlm}, \Othree, correctly identifies the shortest route in only $33$\% of samples, and just $37$\% of its answers successfully connect the map markers---exposing critical limitations in geospatial reasoning and planning.
\end{itemize}
Our task design further reveals a hierarchical, i.e., a foundational or prerequisite, dependence. Specifically, the ability to perform higher-level reasoning depends on the prior availability of low-level map understanding.
In fact, models that struggle with basic symbol identification are unlikely to succeed in measurement or navigation \subtasks. 
Some models exhibit modality-specific strengths, compensating for visual weaknesses through stronger textual-based reasoning.

\dataset provides a comprehensive framework for evaluating and advancing \acp{vlm} in map-based reasoning, offering both a foundation for future extensions and a critical tool for diagnosing current limitations. 
With practical applications in navigation, urban planning, tourism, and real estate, \dataset also lays essential groundwork for building next-generation \acp{vlm} capable of advance geospatial reasoning.
\textbf{Our contributions} are as follows:
\begin{itemize}
    \item We introduce \dataset, the first hierarchically structured benchmark specifically designed to assess \acp{vlm}' \map understanding across three core \tasks: visual recognition, spatial measurement, and navigation.
    \item We evaluate $15$ \acp{vlm}, including cutting-edge proprietary models, and offer detailed analysis of their performance and failure cases.
    \item We demonstrate \dataset’s utility in exposing current model limitations and guiding future improvements in architecture and training for geospatial reasoning.
\end{itemize}


\section{Related works}
This section provides an overview of prior research on \acp{llm} in spatial understanding and geospatial tasks, recent developments in \acp{vlm}, and benchmark datasets for geospatial analysis.

\subsection{LLMs for spatial understanding}
Spatial understanding is crucial for accurate map interpretation. 
Yet, recent studies demonstrate that current language-only \acp{llm} struggle with several tasks and especially relatively simple spatial tasks that would appear trivial for humans. 
Such tasks include 
navigating two-dimensional grids to reach a target~\cite{tang2024grasp},
reasoning about cardinal directions in contextual scenarios~\cite{cohn2024evaluating},
interpreting spatial relationships (visual correspondence, depth perception, object perception, and movement) across multi-frame scene~\cite{xu2025multi}; 
and solving spatial tasks involving complex structures~\cite{yamadaevaluating}, such as hexagonal grids, rings, and trees. 
Such outcomes underscore the inherent challenges of performing spatial reasoning using language-only \acp{llm}. 
This also highlights the potential of incorporating additional modalities, particularly visual inputs, to enhance the geospatial reasoning capabilities of the models.

\subsection{LLMs and LVLMs for geospatial tasks}
Recent research has explored the extent to which \acp{llm} possess implicit geospatial knowledge. 
Some studies have evaluated language-only \acp{llm} on tasks such as \ac{gis} exams~\cite{mooney2023towards} and specific geospatial test cases~\cite{roberts2023gpt4geo}.
These studies reveal that while these models exhibit partial understanding, they also make frequent errors. 
Manvi et al.~\cite{manvigeollm} fine-tuned \emph{GPT-3.5} with custom prompts for estimating population density and found that, despite its limitations, the model displayed a notable level of geospatial awareness. 
Similarly, Bhandari et al.~\cite{bhandari2023large} acknowledged the potential of language-only  \acp{llm} in geospatial applications but emphasized the need for further enhancement. 
Collectively, these works underscore the limitations of language-only \acp{llm} and the need for improved geospatial understanding, e.g., using \acp{mllm}. 

Despite their potential, few \acp{vlm} have been specifically developed for geospatial tasks.
GeoChat, introduced by Kuckreja et al.~\cite{kuckreja2024geochat}, was the first \ac{vlm} to offer multitask conversational capabilities on remote sensing data, covering several tasks such as object recognition and counting.
Yan et al.~\cite{yan2024urbanclip} proposed a model that combines satellite imagery and textual data via contrastive learning to profile urban regions and predict spatial indicators as well as other prompt-guided tasks.
These efforts show that combining visual and textual modalities  are particularly well-suited for advancing cartographic and geospatial understanding.

\subsection{Pre-trained large visual-language models}
\label{sec:sota}
Recent advances in \acp{vlm} have significantly accelerated progress toward general-purpose AI systems capable of processing and reasoning over both visual and textual inputs.

\textbf{Proprietary \acp{vlm}}: The release of ChatGPT~\cite{chatgpt}, demonstrating unprecedented abilities in understanding, reasoning, and generating text across a broad range of tasks, sparked intense interest in developing commercial models even more efficient. 
Leading proprietary \acp{vlm} include \OAI{} \Oone~\cite{jaech2024openai}, \Othree~\cite{o3}, \GeminiP~\cite{gemini2.5}, and \Claude~\cite{claude3.7sonnet}, which demonstrate extended thinking features.
These models are positioned as general-purpose models with ambitions toward \ac{agi}. 

\textbf{Open-source \acp{vlm}}: Open-source \acp{vlm} have greatly influenced the \ac{agi} landscape by democratizing multimodal research---especially, visual and textual data--- enabling rapid innovation and reproducibility. 
Over the past two years, several open-source models have gained prominence, including the 
\emph{LLaVA} series~\cite{liu2023visual, liu2024improved, li2024llavanext-strong, li2024llava}, 
\emph{Llama} series~\cite{llama3.2.11, llama3.2.90, grattafiori2024llama}, 
\emph{MiniGPT}-4~\cite{zhuminigpt}, \emph{VisionLLM}~\cite{wang2024visionllm}, 
\emph{Qwen-VL}~\cite{bai2023qwen, Qwen2.5-VL}, \emph{CogVLM}~\cite{wang2023cogvlm, hong2024cogvlm2}, \emph{InternVL}~\cite{chen2023internvl, chen2024far, chen2024expanding}, and many others~\cite{lin2024vila, alayrac2022flamingo, peng2023kosmos, gao2023llama}.
These models continue to advance the boundaries of visual-language understanding.

However, despite their impressive performance across various benchmarks and tasks, these proprietary and open-source models have not yet been fully evaluated on \ac{cm} interpretation tasks.

\subsection{Benchmark datasets for geospatial tasks}
Most benchmark datasets for geospatial tasks rely heavily on satellite or street-view imagery as visual inputs. 
Lobry et al.~\cite{lobry2020rsvqa} introduced \emph{RSVQA}, a \ac{vqa} dataset based on remote sensing data, targeting tasks such as object presence detection, counting, and classification. 
Yang et al.~\cite{yang2024v} proposed a platform that allows agents to interact with virtual environments constructed from street-view images, enabling the evaluation of \acp{vlm} on tasks like place recognition and \acp{vqa}. 
For urban environments, CityBench~\cite{feng2024citybench} provides a simulator-based evaluation tool, using satellite or street-view images, to assess \acp{llm}' performance on a wide range of urban tasks—from perception and understanding to decision-making. 
In addition, Zhang et al.~\cite{zhang2024good} is conducting ongoing analysis of \emph{GPT-4V}’s capabilities on spatial understanding based on satellite imagery. 
While these studies demonstrate the potential of \acp{vlm} in geospatial tasks, they also expose persistent limitations such as modality dominance (e.g., prioritizing textual over visual inputs) and geospatial biases (e.g., performing better in well-known locations).

To evaluate more directly map-based spatial reasoning, Dihan et al.~\cite{dihan2024mapeval} recently introduced the MapEval dataset, which uses digital map views from \emph{Google Maps}.
Their tasks, framed as multiple-choice questions, center on user-oriented applications such as extracting place information, identifying the nearest \ac{poi} , and routing.
While MapEval tests different input modalities (text, API, visual), it lacks fine-grained, low-level map interpretation tasks, limiting its diagnostic power in revealing foundational weaknesses of \acp{vlm} in \ac{cm} understanding.
Its reliance on multiple-choice formats further restricts insight into model reasoning, and the small number of questions per task may hinder generalizability.

In contrast, \dataset introduces a specifically designed benchmark that spans a hierarchy of tasks---from basic \ac{mapfeat} recognition to advanced reasoning such as scale interpretation and navigation.
Importantly, we introduce a dedicated task for interpreting scale bars and estimating real-world distances---an essential yet underexplored capability for practical geospatial applications.
The open-ended nature of most of our questions better reflects real-world use cases and allows for deeper insight into model behaviors, including hallucinations and reasoning failures.
Moreover, our benchmark evaluates recent proprietary and open-source models, offering a more current and thorough understanding of the state of the field.
A comparison with MapEval is presented in Table~\ref{tab:compare_to_mapeval}.

\begin{figure*}[b]
    \centering
    \includegraphics[width=0.8\linewidth]{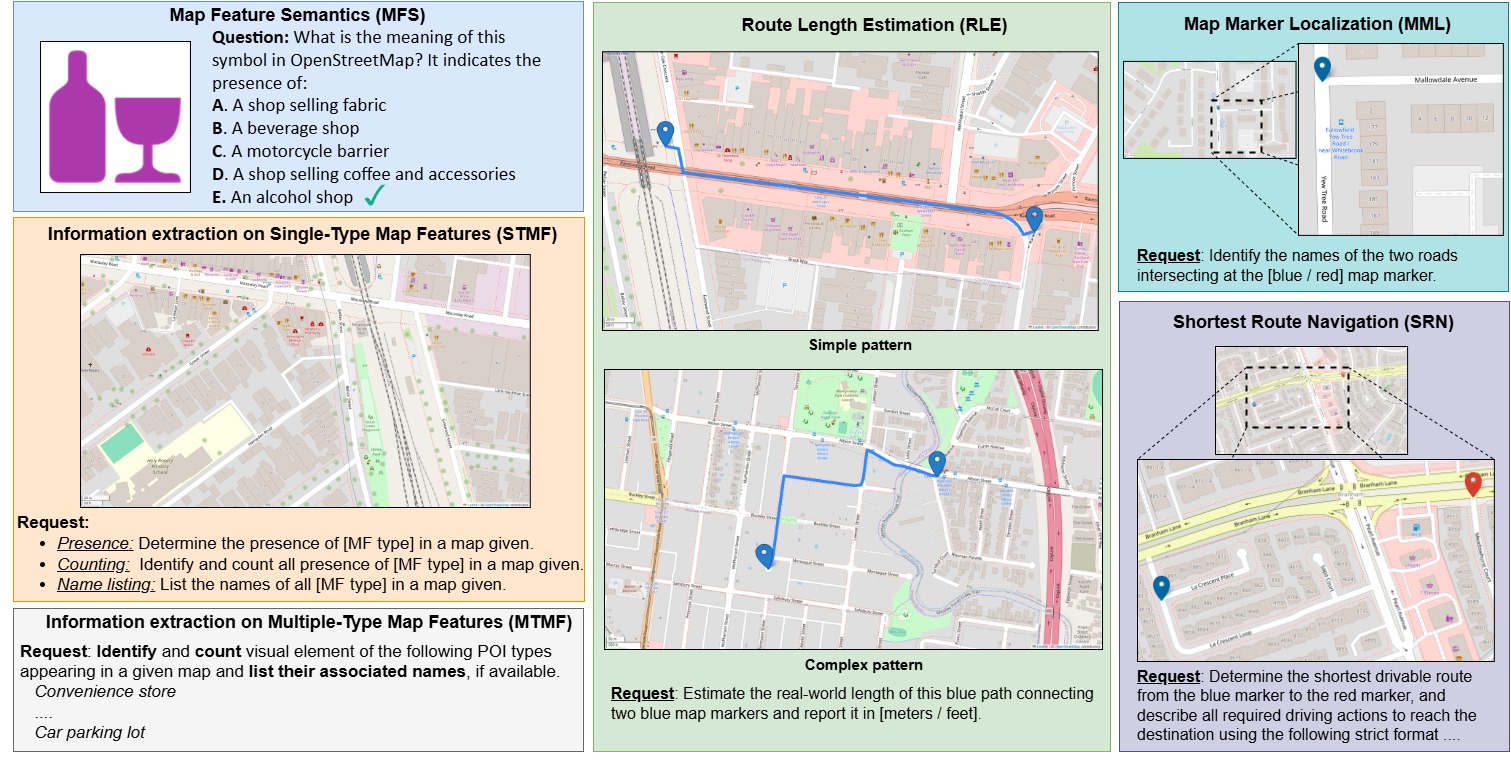}
    \caption{Examples of questions and \questions featured in \dataset. Note: these are short versions of the actual questions and \questions used in the dataset, which can be found in the implementation repository.}
    \Description{A detailed description of the image content.}
    \label{fig:example_bench}
\end{figure*}

\begin{table}[!ht]
    \centering
    \caption{Comparison between \dataset and MapEval. MC stands for multiple-choices. * Include information retrieval testing, while ** evaluate insufficient context detection ability, which are not fully \ac{cm}-related.}
    \resizebox{\columnwidth}{!}{%
    \begin{tabular}{c|ccc|r|c}
        \hline
         &  \multicolumn{3}{c|}{\textbf{Task level}} & \textbf{Number of} & \textbf{Answer} \\
         & \textbf{Low} & \textbf{Mid} & \textbf{High} & \textbf{questions} &  \textbf{format} \\\hline
         \multirow{2}{*}{\textbf{MapEval~\cite{dihan2024mapeval}}} & \multirow{2}{*}{\textcolor{lightgray}{\xmark}} & \textit{Counting}& \textit{Nearby}, \textit{Routing} & \multirow{2}{*}{$700$} & \multirow{2}{*}{MC ($100$\%)}\\ 
         & & \textit{Place Info}*& \textit{Trip}, \textit{Unanswerable}**&  & \\ \hline
         \multirow{2}{*}{\textbf{\dataset (ours)}} &  \multirow{2}{*}{\ac{mfsem}} & \ac{stmf}, \ac{mtmf}, & \ac{rlest}, &  \multirow{2}{*}{$2251$} & Open-ended ($79$\%)\\
         &  & \ac{mmloc} & \ac{srnav} &  & MC($21$\%)\\ \hline
    \end{tabular}
    }
    \label{tab:compare_to_mapeval}
\end{table}


\begin{table}[!ht]
    \centering
    \caption{Key statistics of \dataset.}
    \resizebox{\columnwidth}{!}{%
    \begin{tabular}{lr}
        \hline
         \textbf{Statistics}& \textbf{Number}\\ \hline
         Total questions:& 2251\\ \hline
            *\acs{mfsem} : \acs{stmf} : \acs{mtmf} : \acs{rlest} : \acs{mmloc} : \acs{srnav} & 463 : 510 : 150 : 600 : 250 : 278\\ \hline
 Multiple-choice questions:&\\
 *\acs{mfsem} &463 (21\%)\\ 
 Open-ended questions:                 
&\\
 *\acs{stmf}, \acs{mtmf}, \acs{rlest}, \acs{mmloc}, and \acs{srnav} &1788 (79\%)\\ \hline
         Total OpenStreetMap maps:& 853\\ 
           *San Jose : Manchester : Melbourne& 324 : 297 : 232\\
           *(\acs{stmf} / \acs{mtmf}) : \acs{rlest} : \acs{mmloc} : \acs{srnav} & 150 : 300 : 125 :  278\\ \hline
 Zoom-in levels: (level 19 is the most zoom-in level)&\\
 15 : 16 : 17 : 18 : 19&1 : 7 : 213 : 187 : 445\\ \hline
 Map's size: (width $\times$ height) & 1366 $\times$ 768 pixels\\ \hline
    \end{tabular}
    }
    \label{tab:dataset_stats}
\end{table}

\section{The CartoMapQA dataset}
\subsection{Overview of CartoMapQA}
Our dataset is designed to evaluate \acp{vlm} on three core \tasks:
\begin{inparaenum}[(i)]
    \item map information understanding,
    \item scale and distance interpretation, and
    \item directional reasoning.
\end{inparaenum}
Key statistics are summarized in Table~\ref{tab:dataset_stats} and visual examples are shown in Figure~\ref{fig:example_bench}.

\textbf{\acf{mfsem}:} This \subtask evaluates \acp{vlm}' ability to recognize isolated \acfp{mapfeat}, without contextual cues from a full \map. 
It provides $463$ multiple-choice questions, each tied to a unique \ac{mapfeat} image and offering five candidate answers with a single correct option. 
This \subtask is composed of all \acp{mapfeat} provided in \emph{OpenStreetMap}~\footnote{\url{https://wiki.openstreetmap.org/wiki/OpenStreetMap_Carto/Symbols}}.
\ac{mfsem} tests foundational semantic understanding of cartographic \ac{mapfeat} images.

\textbf{\acf{stmf}:} The goal of this \subtask is to test \acp{vlm}’ ability to detect, classify and analyze a single type of \ac{mapfeat} in full \maps.
We randomly selected three major cities from different countries: San Jose (USA), Manchester (UK), and Melbourne (Australia).
\ac{stmf} includes $150$ \acp{cm} ($50$ per city), all sourced from \emph{OpenStreetMap}. 
Each one is rendered at the finest \textit{zoom-in level} (i.e., $19$) to include finest details, such as \acp{poi}.
To comprehensively evaluate this \subtask, we define three \questions:
\begin{inparaenum}
    \item \texttt{presence}: Determines whether the model can detect and identify at least one instance of a target \ac{mapfeat} type;
    \item \texttt{counting}: Assesses the model’s ability to accurately identify and count all instances of a given \ac{mapfeat} type; and, 
    \item \texttt{name-listing}: Establishes whether the model can correctly extract the textual labels associated with the target \ac{mapfeat} type.
\end{inparaenum}
\ac{stmf} covers $22$ usual \ac{mapfeat} types, including \emph{restaurants}, \emph{convenient stores}, and \emph{parking lots}.
Therefore, each \ac{cm} is associated with several questions, resulting in a total of $510$ questions for this \subtask.
The list of these \acp{mapfeat} is provided in Table~\ref{tab:select_map_features} (Appendix \ref{appendix}).
This framework progressively judges the model's visual detection accuracy, classification, label-association, and basic counting skills. 

\textbf{\acf{mtmf}:} Building on \ac{stmf}, this \subtask examines the \acp{vlm}' \texttt{counting} and \texttt{name-listing} abilities across multiple types of \acp{mapfeat} simultaneously within a single \ac{cm}.
\ac{mtmf} uses the same \acp{cm} as \ac{stmf} and focuses on the same $22$ \ac{mapfeat} types.
This results in a total of $150$ questions for this \subtask.
\ac{mtmf} specifically tests the multitasking capability of \acp{vlm}, establishing a foundation for complex inference tasks that require the ability to interpret and understand information---potentially overlapping---for instance, when analyzing and describing \acp{cm}.

\textbf{\acf{rlest}:} This \subtask evaluates \acp{vlm}’ ability to interpret scale bars and estimate real-world length of predefined routes in meters or feet.
\ac{rlest} includes $300$ \acp{cm} (from the same three cities) rendered at \textit{zoom-in levels} ranging from $15$ to $19$.
As illustrated in Figure~\ref{fig:example_bench}, each \map includes two blue map markers connected by a predefined blue path and a visible scale bar providing a visual indication of distance in both meters and feet. 
Two difficulty levels are defined: 
(1) Simple: Straightforward paths with few turns; 
(2) Complex: Longer paths with multiple turns. 
\ac{rlest} is crucial for assessing scale awareness, a valuable benchmark for advanced \acp{vlm} applications like navigation, planning, or proximity-based services (e.g., nearest \acp{poi}).
The distribution of route lengths is shown in Figure~\ref{fig:lenme_distri} (Appendix~\ref{appendix}).

\textbf{\acf{srnav}}: This \subtask tests \acp{vlm} in terms of directional awareness, finding the shortest route and turn-by-turn navigation.
It is achieved by prompting the model to generate turn-by-turn instructions for the shortest drivable route between two markers on a \ac{cm}: a blue marker (origin) and a red marker (destination).
We generated $278$ valid routes across the three cities (originally $300$; some were filtered out due to readability issues, e.g., overlapping elements), on \acp{cm} rendered at \textit{zoom-in levels} ranging from $17$ to $19$.
For simplicity, we assume the user is seated in a vehicle which is oriented toward the top of the \ac{cm}, at the starting marker.
To standardize outputs, we instruct models to use a structured navigation format: \textit{[blue, <$action_1$>, $road_1$, <$action_2$>, $road_2$, ..., <$action_n$>, $road_n$, red]}, where each \textit{$action_i$} belongs to one of four categories of directional change: \textit{continue straight}, \textit{make a U-turn and continue straight}, \textit{turn left}, and \textit{turn right}.
For instance, if the shortest route involves making a U-turn at the start, followed by driving on ``Main Street'', a left turn onto ``2nd Avenue'', and a right turn onto ``Elm Street'' to reach the destination, the response should be:\
\textit{[blue, make a U-turn and continue straight, Main Street, turn left, 2nd Avenue, turn right, Elm Street, red]}.
This format aligns closely with commercial navigation systems (e.g., \emph{Google Maps}) and supports consistent evaluation.
The distribution of the number of steps (excluding the origin point) in our generated routes is presented in Figure~\ref{fig:step_distribution} (Appendix~\ref{appendix}).
Beyond its central role in navigation systems, \ac{srnav} provides the foundation for more complex applications such as trip planning.

It is worth noting that in both \ac{rlest} and \ac{srnav}, we employ dedicated map markers to designate the origin and destination points---rather than relying on existing \acp{poi}.
This design choice simplify the task and minimize errors that might originate from models' potential inability to detect, classify or associate label with \acp{poi}. 
Nevertheless, this approach requires to specifically evaluate the models' detection capability of such dedicated map markers. 

\textbf{\acf{mmloc}}: This \subtask evaluates \acp{vlm}’ ability to localize a colored map marker within a \map. 
\ac{mmloc} is essential as map markers are commonly utilized to mark specific locations on a \ac{cm}.
Assessing \acp{vlm}' capability on such a task is then of crucial importance.
Rather than asking models for exact coordinates---which can introduce ambiguity---we restrict the position of markers at the intersection of two named roads. 
The model is then prompted to report the names of these intersecting roads. 
Linking marker locations to the identification of road names not only simplifies the annotation process but also aligns directly with the requirements of our \ac{srnav} \subtask.

\begin{figure}[t]
    \centering
    \includegraphics[width=1.0\linewidth]{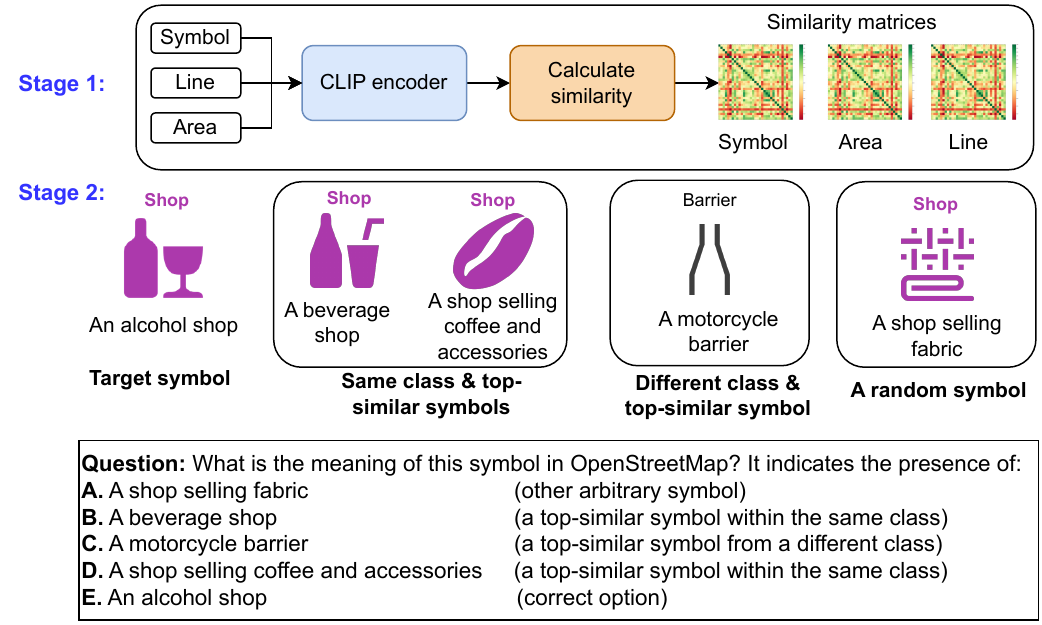}
    \caption{Illustration of the process used to generate answer choices for the multiple-choice questions in the \ac{mfsem} \subtask.}
    \Description{A detailed description of the image content.}
    \label{fig:symsem}
\end{figure}

\subsection{Cartographic map generation}
While well-known platforms such as \emph{Google Maps} and \emph{Apple Maps} provide \maps that are widely used, their strict and restrictive data-sharing policies~\footnote{https://cloud.google.com/maps-platform/terms\#3.-license}~\footnote{https://www.apple.com/legal/internet-services/maps/terms-en.html} limit their suitability for an open dataset.
To ensure that \dataset remains fully accessible to the research community, we adopt \emph{OpenStreetMap} as our \acp{cm} rendering platform.
\Acp{cm} in \dataset are generated using the \emph{Folium} Python package~\cite{folium}, which allows for interactive map rendering and supports built-in \emph{OpenStreetMap} tilesets. 
For \subtasks involving path planning---specifically, \ac{rlest} and \ac{srnav}---we utilize \emph{OSMnx}~\cite{boeing2024modeling} and \emph{NetworkX}~\cite{hagberg2008exploring} packages to compute and visualize paths, including shortest routes, between two points.
These tools enable precise path calculations and ensure faithful representation of \acp{mapfeat}, maintaining spatial and semantic fidelity throughout the dataset.

\subsection{Answer preparation}
\textbf{\ac{mfsem}:} 
To test a model's ability to distinguish the correct answer from distractors and unrelated choices, we design a method leveraging the CLIP encoder~\cite{radford2021learning}, as illustrated in Figure~\ref{fig:symsem}.

First, we compute embeddings for all \ac{mapfeat} images using CLIP and then, generate pairwise Euclidean‐distance similarity matrices separately for each \ac{mapfeat} category: symbol, line and area.
Given that \emph{OpenStreetMap} provides semantic classifications for each \ac{mapfeat} (e.g., shop, amenity, barrier, etc.), we use such classifications alongside the similarity matrices to select distractors based on three criteria:
\begin{inparaenum}
    \item Two most similar \acp{mapfeat} from the same class as the correct answer.
    \item One most similar \ac{mapfeat} from a different class.
    \item One randomly selected \ac{mapfeat}.
\end{inparaenum}
The labels associated with these \acp{mapfeat}, combined with the correct label, form the five choices in each multiple-choice question.
Finally, to prevent positional bias, we randomly shuffle options' order, ensuring that the correct answer appears in a uniformly distributed position.

\textbf{\ac{stmf} and \ac{mtmf}:} We extracted \ac{mapfeat} data within each \ac{cm}’s boundaries directly from \emph{OpenStreetMap}.
To ensure accuracy and consistency between the extracted data and what is visually represented on the rendered \acp{cm}, a human annotator manually verified the counts and names of all \acp{mapfeat}.
This validation step helps eliminate discrepancies that may arise due to rendering limitations, outdated entries, or label mismatches between \emph{OpenStreetMap} data and the actual content displayed in the \acp{cm}.




\textbf{\ac{rlest}, \ac{mmloc} and \ac{srnav}:} Ground-truth for these \subtasks were obtained using \emph{NetworkX} and \emph{OSMnx}, ensuring accurate and consistent spatial reasoning based on real-world geography.
For \ac{rlest}, the true distance was computed using the geographic coordinates (latitude and longitude) of the start and end markers of the predefined path.
For \ac{mmloc}, we randomly selected graph nodes positioned at the intersection of two roads within a selected area of each \ac{cm}.
The names of these intersecting roads were then extracted to form the correct answer.
For \ac{srnav}, we extracted the shortest drivable path as an ordered sequence of graph nodes. 
This sequence was then translated into our standardized turn-by-turn route format by mapping consecutive node pairs to their respective road names and inferring the required directional change. 
Finally, all generated routes were manually verified and corrected to ensure alignment with the visual content of each \ac{cm}.

\begin{table}[!htb]
    \centering
    \fontsize{8pt}{8pt}
    \selectfont
    \caption{Main results of the \acf{mfsem} task. Bold and underlined values indicate the best and second-best performance, respectively, within each group of models.}
    \begin{tabular}{l|rrrr}
    \hline
    \textbf{Model} & \multicolumn{4}{c}{\textbf{Acc.} ($\uparrow$)}\\
    & \textbf{Symbol}& \textbf{Area} & \textbf{Line} & \textbf{Overall}\\ \hline
    \textit{Random baseline} & 0.173 & 0.222 & 0.232 & 0.194\\ \hline
    \llama 11B & 0.507 &  \textbf{0.326}&\underline{0.222}& 0.415\\
    \intern 8B & 0.410 &  0.168 & 0.189 & 0.317\\
    \llava 7B & \textbf{0.655} & 0.221 & \textbf{0.311} & \textbf{0.499}\\
    \qwen  7B & \underline{0.597} & \underline{0.284} & \textbf{0.311} & \underline{0.477}\\ \hline
    \llamaS & 0.590 & \textbf{0.316}&\textbf{0.311}& 0.479\\
    \llama 90B & 0.608 &  \textbf{0.316} &\underline{0.267}& 0.482\\
    \intern 78B&\underline{0.637}&  \underline{0.295}&\underline{0.267}& 0.495\\
    \llava 72B&\textbf{0.683}&  0.263 & 0.233 & \underline{0.510}\\
    \qwen 72B& 0.662 &  \textbf{0.316}&0.256& \textbf{0.512}\\ \hline
 \gptV
& 0.669
&  0.253
&0.300
& 0.512
\\ 
          \gptO
&0.723
&  0.232
&\textbf{0.444}& 0.568
\\
 \oone
&0.777
&  0.368
&0.356
& 0.611
\\
          \othree
& \underline{0.806}&  \underline{0.379}&0.378
& \underline{0.635}
\\ 
 \gemini
&\textbf{0.820}&  \textbf{0.389}&\underline{0.411}& \textbf{0.652}\\
 \claude
&0.687
&  0.305
&0.356
& 0.540
\\ \hline
    \end{tabular}
    
\label{tab:sym}
\end{table}

\begin{table*}[!hb]
    \centering
    \fontsize{8pt}{8pt}\selectfont
    
    \caption{Main results of the \acf{stmf} task. Bold and underlined values indicate the best and second-best performance, respectively, within each group of models.}
    \begin{tabular}{l|rr|rr|rrr}
        \hline
         \textbf{Model} & \multicolumn{2}{c|}{\textbf{Presence}}&\multicolumn{2}{c|}{\textbf{Counting}}& \multicolumn{3}{c}{\textbf{Name listing}}\\ 
           & Acc.  ($\uparrow$)& MF1 ($\uparrow$)& RMSE ($\downarrow$)& $r^2$ ($\uparrow$)& aPrec. ($\uparrow$)& aRec. ($\uparrow$)&aF1 ($\uparrow$)\\ \hline
 \textit{Random baseline}& 0.487& 0.486& 7.278& -5.764
& -& -&-
\\ \hline
 \llama 11B
& 0.597& 0.530& 2.590& 0.143& 0.305& 0.492&0.342
\\
          \intern 8B& \underline{0.707}& \underline{0.706}& 2.662& 0.095& \underline{0.427}& \underline{0.482}&\underline{0.407}\\
          \llava 7B& \textbf{0.754}& \textbf{0.752}& \textbf{1.923}& \textbf{0.528}& 0.365& 0.372&0.340
\\
          \qwen  7B& 0.691& 0.680& \underline{2.270}& \underline{0.342}& \textbf{0.710}& \textbf{0.615}&\textbf{0.624}\\ \hline
          \llamaS & \textbf{0.785}& \textbf{0.784}& 2.398& 0.266& \underline{0.678}& \underline{0.726}&\underline{0.663}\\
 \llama 90B
& \underline{0.749}& \underline{0.746}& 2.322& 0.312& 0.556& 0.595&0.542
\\
 \intern 78B
& 0.733& 0.730& \textbf{1.729}& \textbf{0.618}& 0.544& 0.547&0.519
\\
          \llava 72B& 0.743& 0.736& \underline{1.999}& \underline{0.490}& 0.580& 0.556&0.537
\\
 \qwen 72B
& 0.728& 0.716& 2.166& 0.401& \textbf{0.753}& \textbf{0.744}&\textbf{0.706}\\ \hline
 \gptV
& 0.738& 0.738& 2.240& 0.359& 0.737& 0.695&0.687
\\ 
          \gptO
& 0.780& 0.775& \underline{1.710}& \underline{0.626}& 0.850& 0.839&0.821
\\
 \oone
& 0.785&0.774& 1.812& 0.581& 0.824& 0.829&0.813
\\
          \othree
& \underline{0.801}& \underline{0.797}& 1.789& 0.592& \underline{0.856}& \underline{0.868}&\underline{0.847}\\ 
 \gemini
& \textbf{0.843}& \textbf{0.842}& \textbf{1.367}& \textbf{0.761}& \textbf{0.885}& \textbf{0.929}&\textbf{0.894}\\
 \claude
& 0.801& 0.798& 1.717& 0.624& 0.739& 0.737&0.714
\\ \hline
    \end{tabular}
    
\label{tab:spoi}
\end{table*}

\section{Experiments}
Using \dataset, we evaluate various emerging \acp{vlm}, including both open-source and proprietary models, under a zero-shot setting to assess their ability to understand \maps without prior fine-tuning. 
For each task, we design a tailored prompt which includes output formatting instructions to streamline response processing.
Due to space constraints, full prompt templates are provided in our implementation source code~\footnote{\emph{\url{https://github.com/ungquanghuy-kddi/CartoMapQA.git}}}. 
All experiments were conducted on four NVIDIA A100 GPUs with 80GB of memory.

\subsection{Baselines}
From the \acp{vlm} introduced in subsection~\ref{sec:sota}, we evaluate five recent open-source models that rank among the top performers on the multi-discipline college-level MMMU benchmark~\cite{yue2024mmmu}. 
For each, we include both a lightweight and a large-scale variant:
\begin{inparaenum}
    \item \Intern (8B/78B) ~\cite{chen2024expanding},
    \item \Qwen (7B/72B)~\cite{Qwen2.5-VL},
    \item \Llava (7B/72B)~\cite{li2024llava},
    \item \Llama (11B/90B)~\cite{llama3.2.11, llama3.2.90, dubey2024llama}, and
    \item \LlamaS (109B)~\cite{llama4scout}, a Mixture-of-Experts model with only 17B parameters activated in the inference phase.
\end{inparaenum}

In addition, we test six famous proprietary models: 
\begin{inparaenum}
    \item \emph{GPT-4V}~\cite{gpt4v} (\texttt{gpt-4-turbo-2024-04-09}), 
    \item \GptO~\cite{gpt4o} (\texttt{gpt-4o-2024-08-06}), 
    \item \OAI{} \Oone~\cite{jaech2024openai}(\texttt{o1-2024-12-17}), 
    \item \Othree~\cite{o3} (\texttt{o3-2025-04-16}),
    \item \GeminiP~\cite{gemini2.5} (\texttt{gemini-2.5-pro-preview-03-25}), and 
    \item \Claude~\cite{claude3.7sonnet} (\texttt{claude-3-7-sonnet-20250219}).
\end{inparaenum}
Finally, for \ac{mfsem}, \ac{stmf} and \ac{mtmf} \subtasks, we include a \textit{Random} baseline, which selects answers randomly, to establish a lower-bound reference for model performance.

\subsection{Evaluation process}
We evaluate \ac{vlm} responses using task-specific metrics tailored to each question or \question. 
For multiple-choice questions (\ac{mfsem}), accuracy (Acc.) is used to measure the proportion of correct answers.
\texttt{Presence} \questions (\ac{stmf}) are assessed using both accuracy (Acc.) and macro-average F1-score (MF1) to capture class-wise balance.
Answers to the \texttt{counting} \questions (\ac{stmf}, \ac{mtmf}) are evaluated using both \ac{rmse} and the coefficient of determination ($r^2$).
\texttt{Name-listing} \questions (\ac{stmf}, \ac{mtmf}) are judged based on average precision (aPrec.), average recall (aRec.), and average F1-score (aF1), computed across all \questions.
For the estimation of real-world route length (\ac{rlest}), the estimated length is compared to the ground-truth and evaluated using \ac{rmse}, \ac{mape}, and $r^2$.
For map marker localization (\ac{mmloc}), the ability to correctly identify both road names at the marker’s location is measured with accuracy (Acc.).
For the \ac{srnav} \subtask, we employ the following metrics:
\begin{itemize}
    \item Shortest Route Success Rate (denoted as $SR^2$): Proportion of the route generated by the model that exactly match the ground-truth shortest route.
    \item Average Step Accuracy (denoted as aSA): This metric quantifies how much the route generated by the model overlay with the ground-truth shortest route. 
    For each route, let $L$ be the total number of steps in the true shortest route excluding the starting point and $k$ the length of the longest matching prefix of the model’s generated steps. 
    For example, if the true shortest route is \textit{[blue, Main Street, turn right, 2nd Avenue, red]} and the generated route is \textit{[blue, Main Street, turn right, Main Avenue, red]}, then $L=4$ (total steps without ``blue'') and $k=2$ (only ``Main Street'' and ``turn right'' match the ground truth after the starting point).
    The step accuracy is then defined as $k/L$, and aSA is the mean across all routes.
    \item Connectivity: Proportion of the route generated by the model to form a continuous path from origin to destination, regardless of real‐world traffic constraints (e.g., one‐way streets). 
    This metric evaluates the model’s ability to produce a route connecting two map markers.
\end{itemize}
To streamline the evaluation process, we developed a workflow for extracting key information from model responses.
In cases of invalid or missing answers in multiple-choice or binary \questions, these are replaced with a randomly selected option (similar to~\cite{yue2024mmmu});
with counting \questions, we substitute them with a random number within a predefined range; 
and invalid answers in \texttt{name-listing} \questions are replaced by an empty list. 
However, such fallbacks were rarely needed in our experiments.

\subsection{Main results}
\textbf{\acf{mfsem}:} Table 3 lists the results for this \subtask. 
When considering all \acp{mapfeat} collectively (the 'Overall' column), all models substantially outperform the \textit{Random} baseline, with \GeminiP{} achieving the highest accuracy, closely followed by \Othree.
Among open-source models, \Qwen 72B ranks highest, whereas \Intern 8B scores the lowest.
The performance gap between lightweight and large-scale open-source models is relatively small, but the difference between large-scale open-source and advanced proprietary models is more significant.
When analyzing the performance per \ac{mapfeat} category (symbol, line or area columns), models perform similarly---and generally poorly---on line and area \acp{mapfeat}, indicating limited interpretive ability in these categories.
In contrast, performances on symbol reveal a clearer performance separation between models.
These findings suggest that while current models can, to some extent, interpret \acp{mapfeat} when isolated from any contextual information, substantial room for improvement remains, particularly for open-source models and on \ac{mapfeat} categories requiring more precise understanding.

\begin{table}[!ht]
    \centering
    \caption{Main results of the \acf{mtmf} task. Bold and underlined values indicate the best and second-best performance, respectively, within each group of models.}
    \resizebox{\columnwidth}{!}{%
    \begin{tabular}{l|rr|rrr}
        \hline
         \textbf{Model} &\multicolumn{2}{c|}{\textbf{Counting}}& \multicolumn{3}{c}{\textbf{Name listing}}\\
           & aRMSE ($\downarrow$)& a$r^2$ ($\uparrow$)& amPrec. ($\uparrow$)& amRec. ($\uparrow$)& amF1 ($\uparrow$)\\ \hline
 \textit{Random baseline} & 8.345& -323.447& -& -&-\\ \hline
          \llama 11B
& \textbf{0.919}& \textbf{-0.168}& \textbf{0.315}& 0.217&\underline{0.239}\\
 \intern 8B& 1.285& -6.100& 0.169& \underline{0.414}&0.200
\\
 \llava 7B& 1.131& \underline{-2.290}& 0.126& 0.236&0.133
\\
 \qwen 7B
& \underline{1.032}& -2.408& \underline{0.295}& \textbf{0.614}&\textbf{0.351}\\ \hline
 \llamaS
& 0.833& -2.135& \underline{0.431}& \underline{0.626}&\underline{0.457}\\
 \llama 90B
& 0.968& -2.282& 0.407& 0.419&0.341
\\
 \intern 78B
& \underline{0.825}& -1.575& 0.318& 0.560&0.355
\\
 \llava 72B& 0.826&  \textbf{-0.180}& 0.299& 0.373&0.271
\\
 \qwen  72B
& \textbf{0.686}& \underline{-0.329}& \textbf{0.566}& \textbf{0.655}&\textbf{0.556}\\ \hline
 \gptV
& 0.840& -0.587& 0.434& 0.561& 0.425
\\
 \gptO
& 0.717& 0.058& 0.533& 0.705&0.550
\\
 \oone
& \underline{0.593}& \textbf{0.290}& \textbf{0.691}& 0.797&\textbf{0.712}\\ 
          \othree & 0.612& \underline{0.217}& 0.566& \underline{0.846}& 0.644
\\
 \gemini
& \textbf{0.582}& 0.001&\underline{0.587}& \textbf{0.888}&\underline{0.668}\\
          \claude
& 0.662& -0.844& 0.414& 0.696&0.469
\\ \hline
    \end{tabular}
    }
    
\label{tab:mpoi}
\end{table}

\textbf{\acf{stmf}:} Table~\ref{tab:spoi} reports the results across the three evaluated \questions.
Overall, most models substantially outperform the \textit{Random} baseline.
It is important to note that, for the \texttt{counting} \question, the \textit{Random} baseline selects a number at random between $[0, N]$, with $N=16$ representing the maximum number of a single-type \ac{mapfeat} observed in our dataset.
Proprietary models overall exhibit superior performance, with once again \GeminiP{} achieving the highest performance across all request types. 
Among open-source models, no single model consistently performs the best overall; performance is mostly task-specific. For instance, \Qwen excels for \texttt{name-listing}.

In the \texttt{presence} \question, top models from the different groups (proprietary, lightweight and large-scale open-source) show similar performance, each achieving an MF1 score above $0.75$.
For the \texttt{counting} \question, \GeminiP{} clearly leads with a substantial margin, even over other proprietary models such as \Oone{} and \Othree.
However, as shown in Figure~\ref{fig:models_stmf_count} (Appendix~\ref{appendix}),
even the best-performing models tend to produce accurate counts primarily when the ground-truth is $1$ or $2$, but exhibit a tendency to undercount as the true value increases. 
Finally, for \texttt{name-listing}, \Qwen 7B/72B and \GeminiP{} deliver similar  highest performance among all evaluated models.

\textbf{\acf{mtmf}:} This \subtask asked two \questions simultaneously: \texttt{counting} and \texttt{name-listing}.
To assess overall performance, we calculate metrics separately for all \ac{mapfeat} types within each \ac{cm}, and then average them across all \acp{cm}. 
Therefore, for \texttt{counting}, we report the average \ac{rmse} (aRMSE) and average $r^2$ (a$r^2$); while for \texttt{name-listing}, we use average of micro precision (amPrec.), average of micro recall (amRec.), and average of micro F1-score (amF1).

As shown in Table~\ref{tab:mpoi}, all \acp{vlm} outperform the \textit{Random} baseline.
Among them, \Oone{} ranks higher across most metrics, followed by \Othree{} and \GeminiP. 
Despite this, the overall low scores of all models indicate that \acp{vlm} struggle with this more complex, multi-type reasoning \subtask. 
Additional comparisons in Table~\ref{tab:compare_stmf_mtmf} (Appendix~\ref{appendix})
show that performance on \ac{mtmf} is consistently lower than on the \ac{stmf} \subtask, on both \questions and across most models, confirming that handling multiple \acp{mapfeat} simultaneously is more challenging than addressing them individually. 

\textbf{\acf{mmloc}:} Table~\ref{tab:mml} reports models overall accuracy on this \subtask, along with breakdowns by marker color and zoom-in level. 
\GeminiP{} again exhibits superior performance with a large margin over all other models.
Among open-source options, \Qwen 7B and 72B achieve the highest accuracy in the lightweight and large-scale groups, respectively.
Performance is consistent across marker colors, suggesting that color variation does not significantly impact model detection ability.
However, accuracy increases with higher zoom-in level (i.e., more focused views), suggesting that reduced information facilitates marker detection.
Despite these trends, all models still exhibit relatively low overall accuracy, underscoring the crucial need for improvements before addressing more demanding \subtasks that rely on accurate marker localization---such as the \ac{srnav} \subtask.

\textbf{\acf{rlest}:} This \subtask requires \acp{vlm} to perform multiple intermediate reasoning steps, ultimately estimating real-world distances.
Note that each request specified the unit for length measurement: either meters or feet.
To support this process, we employed a \ac{cot}~\cite{wei2022chain} prompting strategy, guiding models through step-by-step reasoning for route length estimation. 
Due to space constraints, the detailed CoT prompt is available in our source code.
Table~\ref{tab:task2_cot} (Appendix~\ref{appendix}) demonstrates that
such an enhancement improves performance across most models and metrics compared to a non-\ac{cot} strategy.

Table~\ref{table:task2} lists the complete results of this \subtask using the \ac{cot} prompt, with \Oone{} and \Othree{} achieving the top two scores across most cases and metrics.
For open-source models, \Qwen 72B delivered the best results, but it frequently repeats specific numbers regardless of the provided \ac{cm} (as illustrated in Figure~\ref{fig:qwen_lenmes} from Appendix~\ref{appendix}).
This behavior likely stems from its difficulty in detecting map markers (as evidenced in Table~\ref{tab:mml}), leading to failures in identifying the blue path and, consequently, generating arbitrary length estimates.
Despite improvements with \ac{cot} prompting, error rates remain high across all models, especially open-source ones.
Such an observation further highlights the challenge associated to these \subtasks and the need for further advancement.

\begin{table}[!ht]
    \centering
    \caption{Main results of the \acf{mmloc} task. Bold and underlined values indicate the best and second-best performance, respectively, within each group of models.}
    \resizebox{\columnwidth}{!}{%
    \begin{tabular}{l|rr|rrr|r}
        \hline
         \textbf{Model} &\multicolumn{6}{c}{\textbf{Acc.} ($\uparrow$)}\\
 & \multicolumn{2}{c|}{\textbf{Marker color}}& \multicolumn{3}{c|}{\textbf{Zoom-in level}}&\textbf{Overall}\\
           & Blue & Red & 17& 18&  19&\\ \hline
          \llama 11B
& \underline{0.128}& \underline{0.128}& \textbf{0.106}& \underline{0.038}& \underline{0.263}&\underline{0.128}\\
 \intern 8B& 0.112& 0.056
& 0.000& 0.029& 0.225
&0.084
\\
 \llava 7B& 0.048& 0.048
& 0.000& 0.019& 0.125
&0.048
\\
 \qwen 7B
& \textbf{0.248}& \textbf{0.240}& \underline{0.096}& \textbf{0.152}& \textbf{0.513}&\textbf{0.244}\\ \hline
 \llamaS
& \underline{0.260}& \underline{0.264}& 0.029& \textbf{0.333}& \underline{0.500}&\underline{0.260}\\
 \llama 90B
& 0.136& 0.152
& 0.038& 0.091& 0.325
&0.144
\\
 \intern 78B
& 0.112& 0.176
& \underline{0.048}& 0.106& 0.300
&0.144
\\
 \llava 72B& 0.080&  0.088
& 0.000& 0.091& 0.188
&0.084
\\
 \qwen 72B
& \textbf{0.288}& \textbf{0.304}& \textbf{0.197}& \underline{0.173}& \textbf{0.538}&\textbf{0.296}\\ \hline
 \gptV
& 0.064& 0.096
& 0.077& 0.030&  0.125
&0.080
\\
 \gptO
& 0.344& 0.360
& 0.240& 0.242& 0.588
&0.352
\\
 \oone
& \underline{0.416}& \underline{0.432}& \underline{0.308}& \underline{0.379}& \underline{0.613}&\underline{0.424}\\ 
          \othree
& 0.304& 0.304
& 0.212& 0.227&  \underline{0.613}&0.344
\\
 \gemini
& \textbf{0.608}& \textbf{0.592}&\textbf{0.471}& \textbf{0.636}& \textbf{0.738}&\textbf{0.600}\\
          \claude
& 0.224& 0.240
& 0.135& 0.227& 0.363
&0.232
\\ \hline
    \end{tabular}
    }
    
\label{tab:mml}
\end{table}

\textbf{\acf{srnav}:} Evaluation results for this \subtask are presented in Table~\ref{tab:navigation}.
It includes both the overall scores and breakdowns by zoom-in level.
As with previous \subtasks, proprietary models outperform open-source ones, with \Othree{} achieving the highest overall performance, followed by \Oone.
However, \Othree{} performance remains modest with only $33.8$\% of proposed routes matching the shortest routes ($SR^2$ ) and $37.8$\% of the routes that correctly connect both starting and destination markers (connectivity).
Open-source models perform significantly worse, with their generated routes rarely matching the true shortest path.
While inaccurate marker localization is likely a contributing factor, this does not fully explain the poor performance.
In fact, \GeminiP{}, which achieves the highest accuracy ($0.6$) in the \ac{mmloc} \subtask, only attains low $SR^2$ ($0.184$) and connectivity ($0.273$) scores.
This suggests that other factors, such as limited geospatial reasoning abilities, may also be at play.

As observed in the \ac{mmloc}, performance improves with higher zoom-in level, one more time suggesting that more focused map views---with reduced contextual information---facilitate \subtasks resolution.
Altogether, these findings again highlight a substantial gap in current \acp{vlm} capabilities to perform this \subtask accordingly and more generally complex geospatial reasoning.
They also emphasize the need to expand \dataset in the future with additional \subtasks to better uncover and address the root causes of these limitations.

\begin{table*}[!hb]
    \fontsize{9pt}{9pt}\selectfont
    \centering
    \caption{Main results of the \acf{rlest} task. Bold and underlined values indicate the best and second-best performance, respectively, within each group of models.}
    \resizebox{\linewidth}{!}{%
    \begin{tabular}{l|r|r|r|r|r|r|r|r|r|r|r|r}
        \hline
         \textbf{Model} & \multicolumn{6}{c|}{\textbf{Simple}}& \multicolumn{6}{c}{\textbf{Complex}}\\ 
           & \multicolumn{3}{c|}{\textbf{Meter}}& \multicolumn{3}{c|}{\textbf{Feet}}& \multicolumn{3}{c|}{\textbf{Meter}}& \multicolumn{3}{c}{\textbf{Feet}}\\ 
           & RMSE ($\downarrow$)& MAPE ($\downarrow$)& $r^2$ ($\uparrow$)& RMSE ($\downarrow$)& MAPE ($\downarrow$)& $r^2$ ($\uparrow$)& RMSE ($\downarrow$)& MAPE ($\downarrow$)& $r^2$ ($\uparrow$)& RMSE ($\downarrow$)& MAPE ($\downarrow$)& $r^2$ ($\uparrow$)\\ \hline
 \llama 11B
& 1446.46& 16.80& -44.91& 1474.81& 4.83& -3.43& 1651.13& 4.91& -6.88& \underline{2023.44}& 1.02& \underline{-0.10}\\
          \intern 8B& 601.30& \underline{7.66}& -6.93& 1416.15& \underline{2.85}& -3.09& 804.98& \underline{2.15}& -0.87& 2125.96& \underline{0.91}& -0.21
\\
          \llava 7B& \underline{586.83}& 11.79& \underline{-6.56}& \textbf{711.12}& 3.10& \textbf{-0.03}& \textbf{600.19}& 2.21& \textbf{-0.04}& 2288.64& \textbf{0.76}& -0.41
\\
          \qwen 7B& \textbf{543.59}& \textbf{4.29}& \textbf{-5.48}& \underline{714.44}& \textbf{2.26}& \underline{-0.04}& \underline{647.80}& \textbf{2.05}& \underline{-0.21}& \textbf{1852.24}& 0.97& \textbf{0.08}\\ \hline
 \llamaS & \underline{250.32}& \underline{1.44}& \underline{-0.37}& 1837.56& \underline{1.34}& -5.88& 518.02& \underline{0.67}& 0.22& 2251.80& 0.72&-0.36
\\
 \llama 90B
& 318.40& 5.48& -1.22& 655.49& 2.81& 0.12& 562.66& 1.26& 0.08& \textbf{1499.69}& 0.63&\textbf{0.40}\\
 InternVL 2.5 78B& 363.22& 3.55& -1.89& \underline{618.14}& 1.44& \underline{0.22}& \underline{509.27}& 1.06& \underline{0.25}& \underline{1911.23}& \underline{0.61}&\underline{0.02}\\
 LlaVA-OV 72B& 530.95& 3.62& -5.19& 671.92& 1.84& 0.08& 626.88& 1.71& -0.14& 1978.15& 0.64&-0.05
\\
          \qwen 72B
& \textbf{170.43}& \textbf{1.31}& \textbf{0.36}& \textbf{471.57}& \textbf{0.58}& \textbf{0.55}& \textbf{433.74}& \textbf{0.46}& \textbf{0.46}& 1914.02& \textbf{0.50}& \underline{0.02}\\ \hline
 \gptV
& 271.15& 0.97& -0.61& 549.96& 0.86& 0.38& 449.77&0.48& 0.42& 1678.36& 0.53& 0.24
\\ 
          \gptO
& 208.86& 0.93& 0.04& 526.32& 0.51& 0.44& \underline{314.93}& 0.42&\underline{0.71}& \underline{1344.83}& 0.42& \underline{0.51}\\
 \oone
& \underline{104.25}& \textbf{0.52}& \underline{0.76}& \textbf{290.88}& \underline{0.43}& \textbf{0.83}& \textbf{305.12}& \textbf{0.32}& \textbf{0.73}& \textbf{1096.74}& \underline{0.40}&\textbf{0.68}\\
 \othree
& \textbf{91.69}& 0.70& \textbf{0.82}& \underline{476.21}& 0.55& \underline{0.54}& 386.96& \underline{0.33}& 0.57& 1519.38& \textbf{0.36}&0.38
\\
 \gemini
&127.25&\underline{0.53}& 0.64& 485.77& \textbf{0.42}& 0.52&458.06& 0.43& 0.39& 1856.34& 0.55&0.07
\\
          \claude
& 130.64& 0.82& 0.63& 506.15& 0.44& 0.48& 454.79& 0.44& 0.40& 1899.86& 0.55& 0.03
\\ \hline
    \end{tabular}
    }
\label{table:task2}
\end{table*}

\begin{figure*}
    \begin{subfigure}{.4\textwidth}
      \centering
      \includegraphics[width=1.0\linewidth]{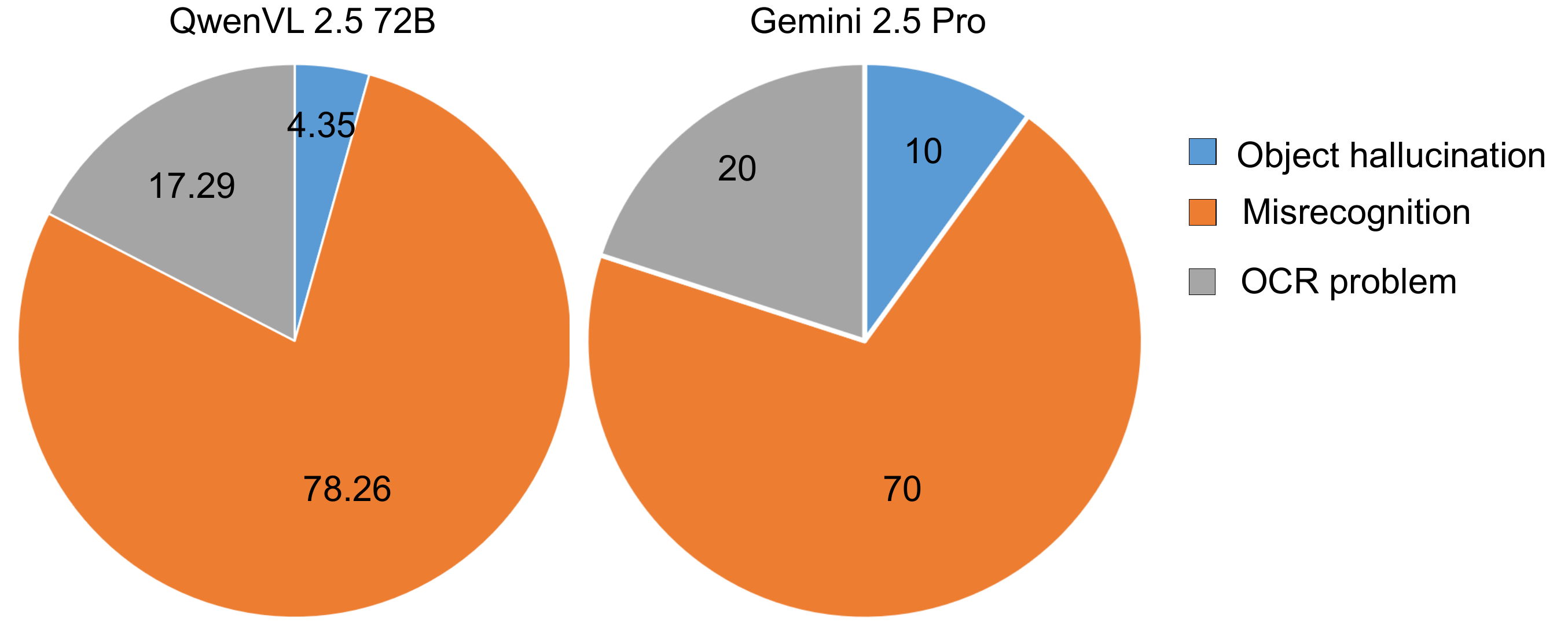}
      \caption{Distribution of main issues.}
      \label{fig:pie_chart}
    \end{subfigure}
    \begin{subfigure}{.5\textwidth}
      \centering
      \includegraphics[width=1.0\linewidth]{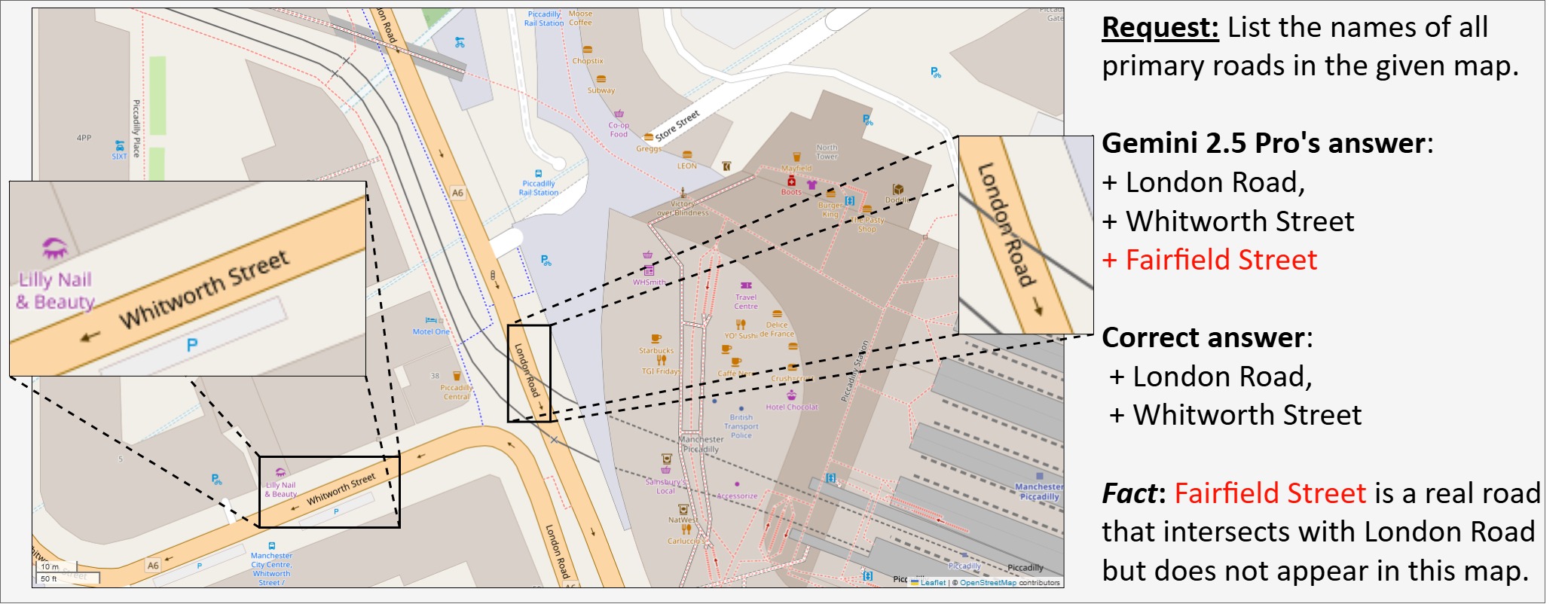}
      \caption{Example of object hallucination.}
      \label{fig:hallu}
    \end{subfigure} \\
    
    \begin{subfigure}{.5\textwidth}
      \centering
      \includegraphics[width=0.96\linewidth]{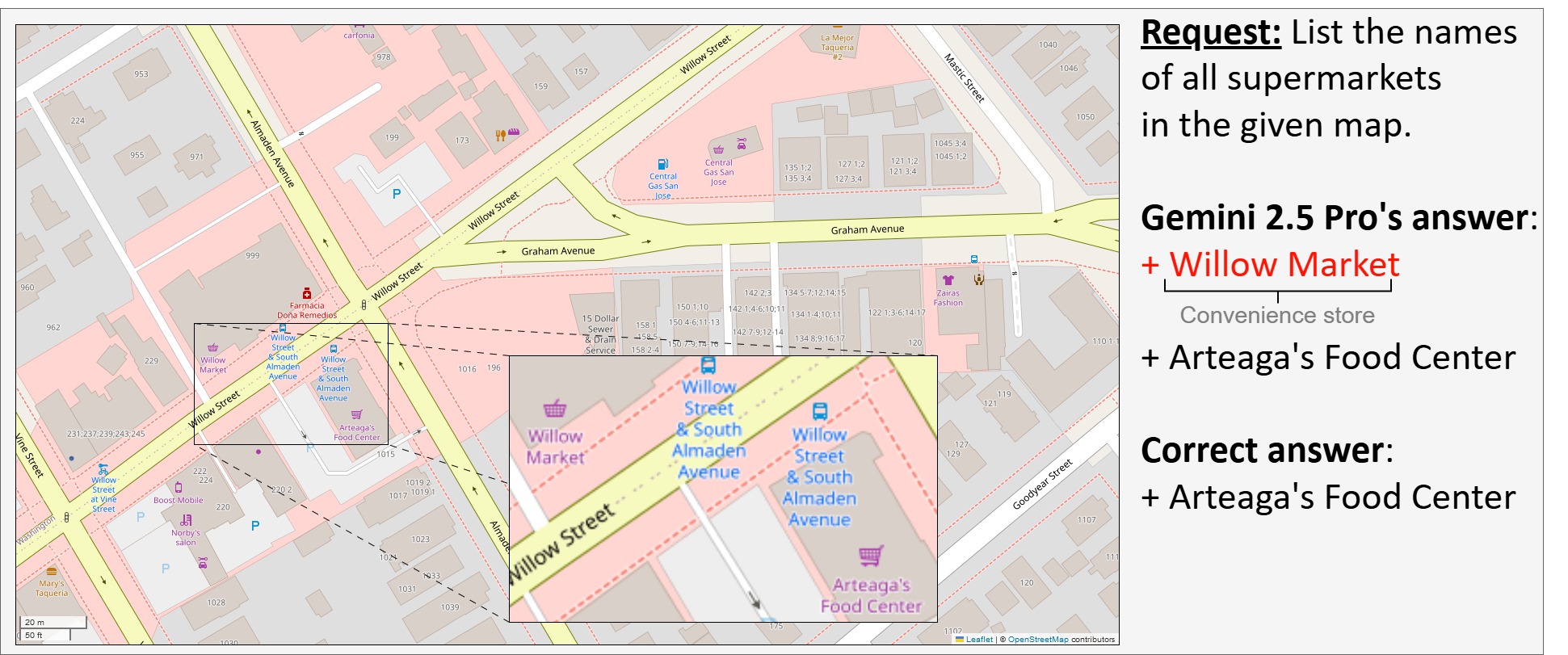}
      \caption{Example of misrecognition.}
      \label{fig:mc}
    \end{subfigure}
    \begin{subfigure}{.49\textwidth}
      \centering
      \includegraphics[width=1.0\linewidth]{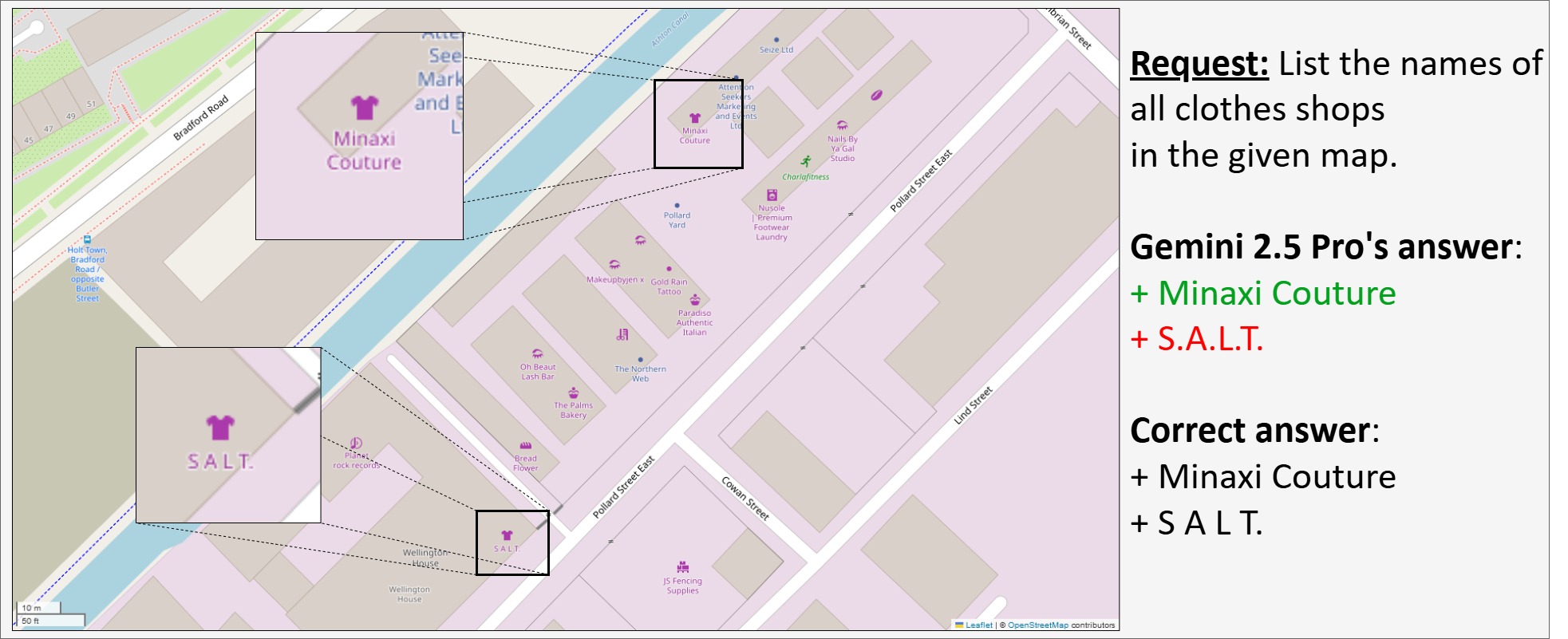}
      \caption{Example of OCR problems.}
      \label{fig:ocr_repeat}
    \end{subfigure}
    \caption{Distribution and examples of the main issues identified in $50$\% of the incorrect answers from \GeminiP{} and \Qwen 72B in the \acf{stmf} \subtask with \texttt{name-listing} \questions. Note: due space constraints, requests shown are short versions of the actual ones.}
    \Description{A detailed description of the image content.}
    \label{fig:poi_ana}
\end{figure*}

\begin{table*}[!hb]
    \centering
    \fontsize{9pt}{9pt}\selectfont
    \caption{Main results of the \acf{srnav} task. Bold and underlined values indicate the best and second-best performance, respectively, within each group of models.}
    \resizebox{\linewidth}{!}{%
    \begin{tabular}{l|r|r|r|r|r|r|r|r|r|r|r|r}
        \hline
         \textbf{Model} & \multicolumn{9}{c|}{\textbf{Zoom-in level}}&\multicolumn{3}{c}{\textbf{Overall}}\\ 
           & \multicolumn{3}{c|}{\textbf{Zoom 17}}& \multicolumn{3}{c|}{\textbf{Zoom 18}}& \multicolumn{3}{c|}{\textbf{Zoom 19}}& \multicolumn{3}{c}{}\\ 
           & $SR^2$ ($\uparrow$)& aSA ($\uparrow$)& Connectivity ($\uparrow$)& $SR^2$ ($\uparrow$)& aSA($\uparrow$)& Connectivity ($\uparrow$)& $SR^2$ ($\uparrow$)& aSA($\uparrow$)& Connectivity ($\uparrow$)& $SR^2$ ($\uparrow$)& aSA($\uparrow$)& Connectivity ($\uparrow$)\\ \hline
 \llama 11B
& 0.000& 0.047& 0.000& 0.000& \underline{0.078}& \textbf{0.011}& 0.000& \underline{0.111}& \textbf{0.024}& 0.000& \underline{0.077}& \textbf{0.011}\\
          \intern 8B& 0.000& 0.030& 0.000& 0.000& 0.059& 0.000& 0.000& 0.046& \underline{0.012}& 0.000& 0.045& \underline{0.004}\\
          \llava 7B& 0.000& \underline{0.071}& 0.000& 0.000& 0.062& 0.000& 0.000& 0.038& \underline{0.012}& 0.000& 0.058& \underline{0.004}\\
          \qwen 7B& 0.000& \textbf{0.094}& 0.000& 0.000& \textbf{0.126}& 0.000& 0.000& \textbf{0.086}& 0.000& 0.000& \textbf{0.102}& 0.000
\\ \hline
 \llamaS & 0.000& 0.028& 0.000& 0.000& 0.122& 0.011& \underline{0.024}& 0.119& \underline{0.024}& \underline{0.007}& 0.086&0.011
\\
 \llama 90B
& 0.000& 0.096& 0.000& 0.000& 0.151& 0.011& 0.000& 0.115& 0.012& 0.000& 0.119&0.007
\\
 \intern 78B
& 0.000& 0.064& 0.000& 0.000& 0.097& 0.000& 0.000& 0.105& \underline{0.024}& 0.000& 0.087&0.007
\\
 \llava 72B& \textbf{0.010}& \underline{0.087}& \textbf{0.010}
& \textbf{0.011}& \underline{0.172}& \textbf{0.034}& 0.000& \underline{0.154}& 0.012& \underline{0.007}& \underline{0.135}&\underline{0.018}\\
          \qwen 72B
& 0.000& \textbf{0.099}& \textbf{0.010}& 0.000& \textbf{0.228}& \underline{0.022}& \textbf{0.036}& \textbf{0.205}& \textbf{0.060}& \textbf{0.011}& \textbf{0.173}& \textbf{0.029}
\\ \hline
 \gptV
& 0.000& 0.023& 0.029& 0.000& 0.069& 0.045& 0.024&0.082& 0.107& 0.007& 0.056& 0.058
\\ 
          \gptO
& 0.010& 0.084& 0.019& 0.022& 0.135& 0.101& 0.060& 0.184&0.107& 0.029& 0.131& 0.072
\\
 \oone
& \underline{0.133}& \underline{0.278}& \underline{0.200}& \underline{0.281}& \underline{0.419}& \underline{0.404}& \underline{0.464}& \underline{0.419}& \underline{0.500}& \underline{0.281}& \underline{0.423}& \underline{0.356}
\\
 \othree
& \textbf{0.238}& \textbf{0.352}& \textbf{0.267}& \textbf{0.292}& \textbf{0.448}& \textbf{0.326}& \textbf{0.512}& \textbf{0.665}& \textbf{0.571}& \textbf{0.338}& \textbf{0.477}& \textbf{0.378}
\\
 \gemini
&0.067&0.167& 0.124& 0.157& 0.307& 0.258&0.357& 0.307& 0.476& 0.184& 0.311&0.273
\\
          \claude
& 0.105& 0.188& 0.124& 0.101& 0.209& 0.191& 0.155& 0.241& 0.238& 0.119& 0.211& 0.180
\\ \hline
    \end{tabular}
    }
    
\label{tab:navigation}
\end{table*}

\begin{figure}[t]
    \begin{subfigure}{.25\textwidth}
    \centering
    \includegraphics[width=1\linewidth]{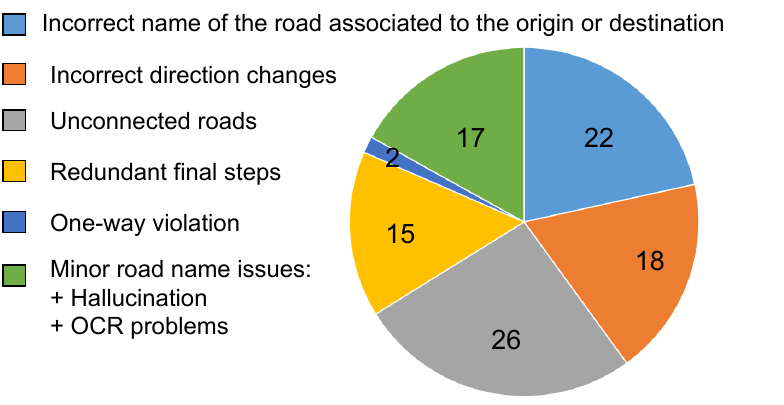}
    \caption{Distribution of main issues found from 30\% of the incorrect routes.}
    \label{fig:navi_ana}
    \end{subfigure}
    \begin{subfigure}{.19\textwidth}
    \centering
    \includegraphics[width=1.15\linewidth]{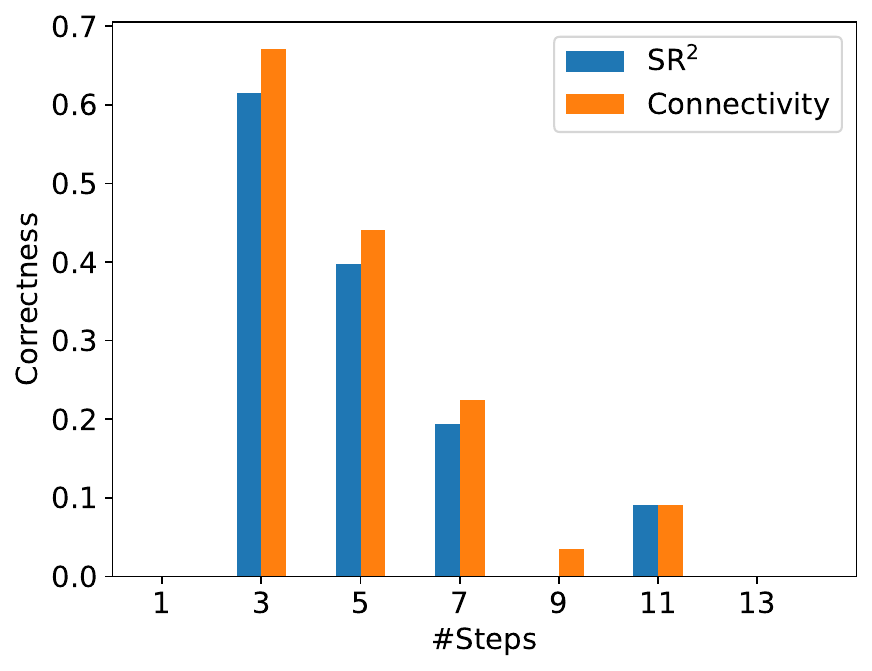}
    \caption{Comparison of correctness ($SR^2$ and connectivity) with varying number of steps in the shortest route.}
    \label{fig:navi_numstep_vs_correctness}
    \end{subfigure}
    \caption{Analysis from the \emph{o3} model in the \acf{srnav} \subtask.}
    \Description{A detailed description of the image content.}
\end{figure}

\subsection{Analysis and discussions}
This section provides a deeper analysis of \acp{vlm}' performance across all \tasks and discusses key insights and potential directions for enhancing their understanding of \ac{cm}.

\textbf{\acl{mapfeat} Understanding:} We examined $50\%$ of incorrect answers on the \texttt{name-listing} \question in the \ac{stmf} \subtask, from \Qwen 72B and \GeminiP---the top-performing open-source and proprietary models, respectively.
Such an analysis revealed three main issues: object hallucination,  incorrect association of \ac{mapfeat} type (misrecognition), and \ac{ocr} errors.
Figure~\ref{fig:poi_ana} provides examples of each issue and their occurrence across both models.
Hallucination was rare, with only one or two occurrences per model, indicating that both models generally base their answers on visible content in the \ac{cm}.
However, misrecognition and \ac{ocr} errors were far more prevalent.
In the case of misrecognition, while \acp{vlm} can detect \ac{poi} and read associated names on \acp{cm}, they often struggle to assign them to the correct \ac{mapfeat} types---frequently confusing closely related types such as restaurants, fast-food stores, and coffee shops (see Figure~\ref{fig:confus_o1} from Appendix~\ref{appendix}).

These findings highlight two critical limitations:
\begin{inparaenum}
    \item current \ac{ocr} capabilities are not sufficiently robust for cartographic \subtasks, and
    \item semantic reasoning for visual entities remains shallow, particularly when distinguishing between conceptually adjacent types.
\end{inparaenum}
A promising direction for future work is to enhance the vision encoder, such as using a high-resolution or map-specific encoder.

\textbf{Map Scale Understanding:} We analyzed $50$ randomly selected responses from \Oone{} on the \ac{rlest} \subtask, which demonstrated the highest performances across most cases and metrics.
This review revealed one key strength: the model demonstrates a solid ability to read the scale bar, accurately extracting its numerical value and associated unit, in $84\%$ of cases.
An illustrative example is shown in Figure~\ref{fig:example_merit} from Appendix~\ref{appendix}.
However, this analysis also unveiled an unexpected behavior: in $8\%$ of these responses, the model attempted to interpret the scale unit into inches or centimeters---units irrelevant to the context of \acp{cm}.
This suggests occasional confusion possibly stemming from the data used during training.

While \Oone{} shows promising capabilities, its non-negligible error rate demonstrate that models could be improved.
Future work could investigate specialized pretraining or alignment strategies to better capture scale, numeric value and unit as well as estimate distance.

\textbf{Map Navigation:} We analyzed $30$\% of incorrect routes (randomly selected) provided by the \Othree{} model---the top performer on this task---to understand the sources of failure.
Figure~\ref{fig:navi_ana} presents the distribution of six most frequent error types observed:
\begin{inparaenum}[(i)]
    \item Incorrect name of the road associated to the origin or destination ($22$\%): These issues could stem from inaccurate localization of map markers or misidentification of road names.
    \item Incorrect direction changes ($18$\%): The model frequently failed to accurately provide turning directions, likely reflecting limitations in geospatial reasoning or misinterpretation of road layouts.
    \item Unconnected roads ($26$\%): The most prevalent issue, where the model suggested paths that involve transitions between roads that are not physically connected, indicating gaps in topological map understanding.
    \item One-way violations ($2$\%): Although rare, these errors---where the model proposed driving against one-way traffic---are critical. They may result from an inability to detect one-way indicators (e.g., arrows), and pose significant safety risks in real-world navigation tasks.
    \item Redundant final steps ($15$\%): The model occasionally added additional steps after reaching the destination, causing unnecessary detours. This suggest a lack of clear identification of the destination or routing capabilities.
    \item Minor road name issues ($17$\%): These include hallucinated or incorrectly read road names, again demonstrating \ac{ocr} limitations.
\end{inparaenum}
Figure~\ref{fig:example_navi} and~\ref{fig:example_navi_success} in Appendix~\ref{appendix} illustrate examples of these failure types and of correctly generated routes, respectively.

To further assess model performance,  Figure~\ref{fig:navi_numstep_vs_correctness} examines \Othree{} scores based on the number of steps in the shortest route.
Notably, the model achieves $SR^2$ scores over $0.4$ for short routes requiring only three to five steps (i.e., one or two turns), but performance rapidly declines for longer, more complex routes, highlighting a key limitation.
These findings underscore that while \Othree{} exhibits some geospatial reasoning capabilities, it might lack robust topological awareness, road segmentation capability and context-sensitive route planning.
Future research might explore hybrid approaches combining \acp{vlm} with graph-based map parsers, or strategies to gradually build up complex navigation skills.

\section{Conclusion and future work} 
In this paper, we introduced \dataset, a benchmark dataset designed to assess the capabilities of \acfp{vlm} in understanding \map. 
\dataset tests a broad range of skills---i.e., map features, and their associated text, recognition and extraction, scale interpretation, directional understanding, and turn-by-turn navigation---essential for comprehensive \acf{cm} comprehension and geospatial reasoning.
Our evaluations reveal that, despite recent progress, current \acp{vlm} still face significant challenges in reasoning over \acp{cm}, particularly in handling complex geospatial tasks.
By structuring \dataset into a hierarchy of \subtasks with increasing complexity, we provide a rigorous and valuable benchmark that not only help identifying specific weaknesses but also supports the development of more robust, map-aware \acp{vlm}. 


The insights gained from our study can inform future efforts on how to fine-tune or design \ac{vlm} architectures tailored to the challenges of cartographic reasoning.
Future directions include expanding \dataset with additional \subtasks to expose more nuanced limitations, and exploring advanced capabilities such as complex navigation and social indicator estimation.

\bibliographystyle{ACM-Reference-Format}
\bibliography{main}

\appendix
\clearpage
\section{Appendix} 
\label{appendix}

\begin{figure}[ht]
    \centering
    \includegraphics[width=0.8\linewidth]{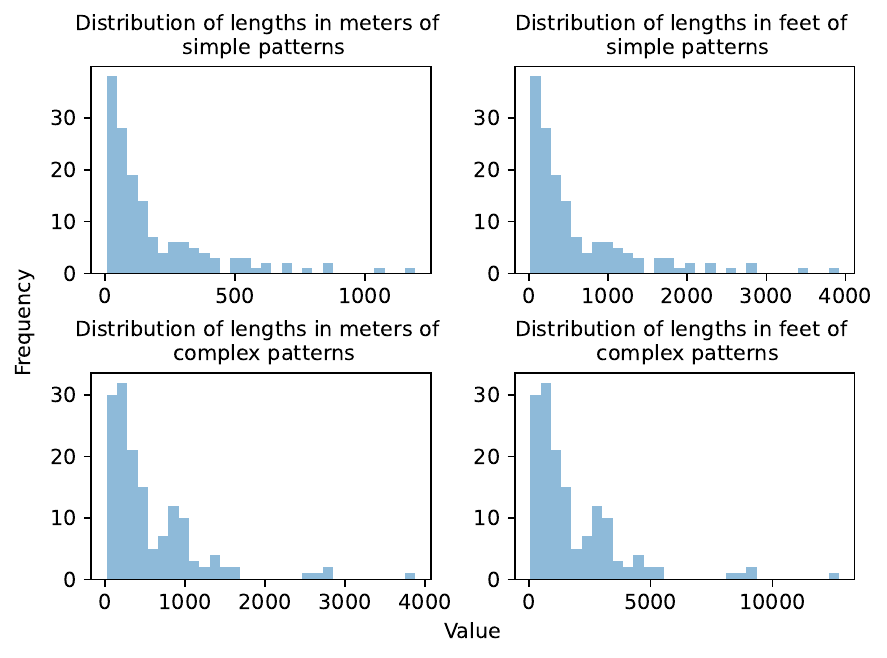}
    \caption{Distribution of route lengths in the \acf{rlest} task.}
    \Description{A detailed description of the image content.}
    \label{fig:lenme_distri}
\end{figure}

\begin{figure}[ht]
    \centering
    \includegraphics[width=0.5\linewidth]{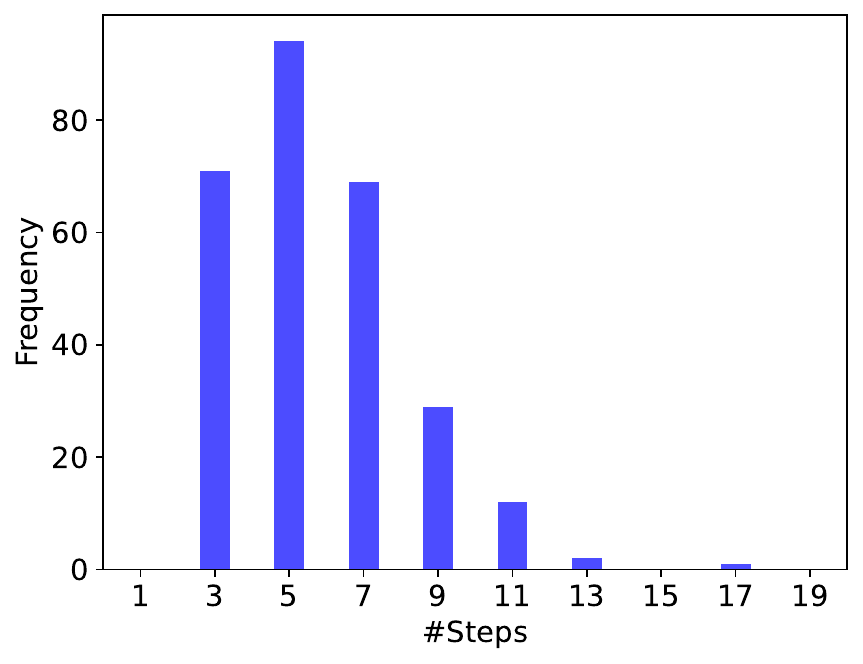}
    \caption{Distribution of the number of steps in the \acf{srnav} \subtask.}
    \Description{A detailed description of the image content.}
    \label{fig:step_distribution}
\end{figure}

\begin{figure}[ht]
    \begin{subfigure}{.23\textwidth}
        \centering
        \includegraphics[width=0.8\linewidth]{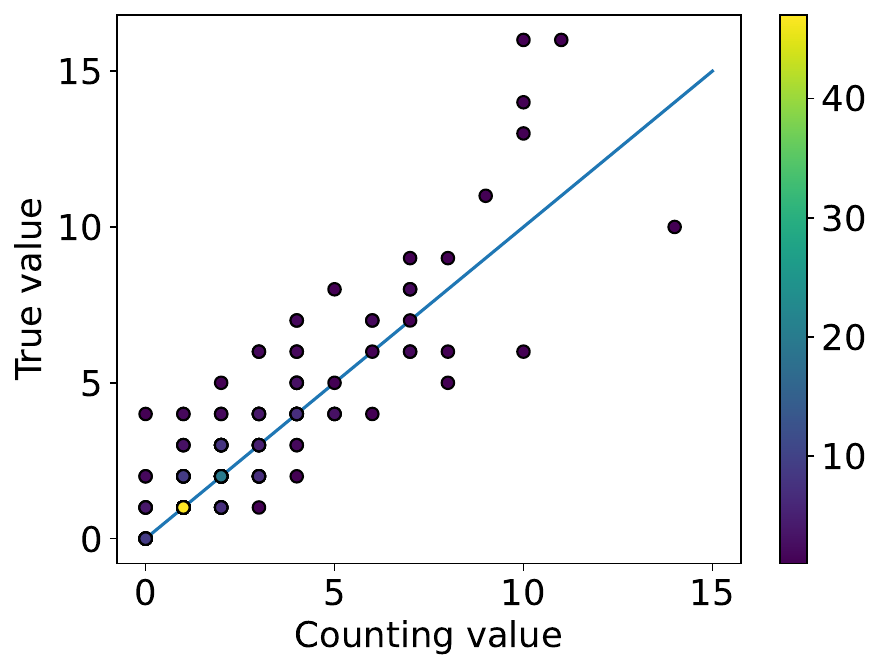}
        \caption{\gemini}
        \label{fig:gemini_stmf_count}
    \end{subfigure}
    \begin{subfigure}{.23\textwidth}
        \centering
        \includegraphics[width=0.8\linewidth]{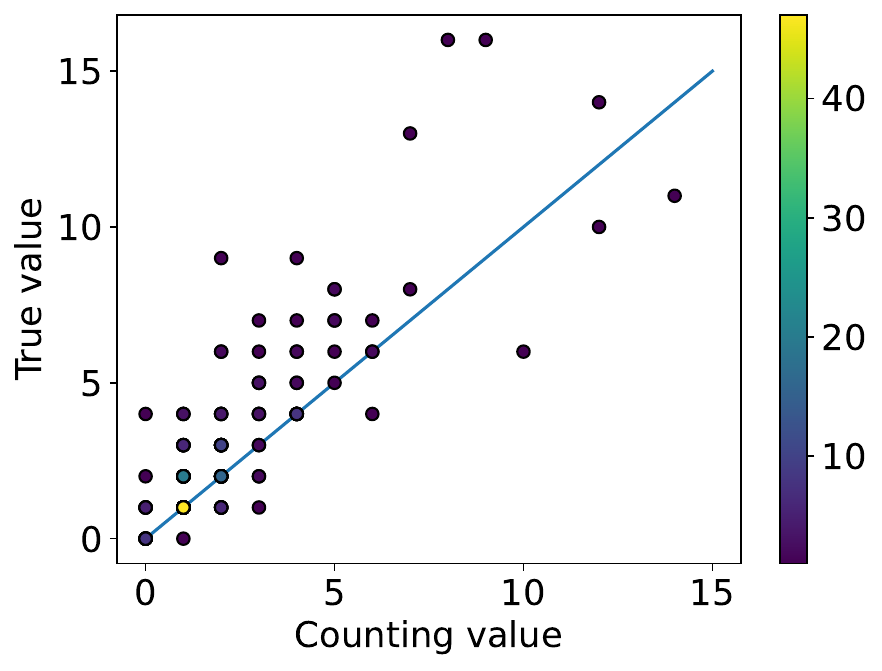}
        \caption{\gptO}
        \label{fig:gpt4o_stmf_count}
    \end{subfigure}
    \begin{subfigure}{.23\textwidth}
        \centering
        \includegraphics[width=0.8\linewidth]{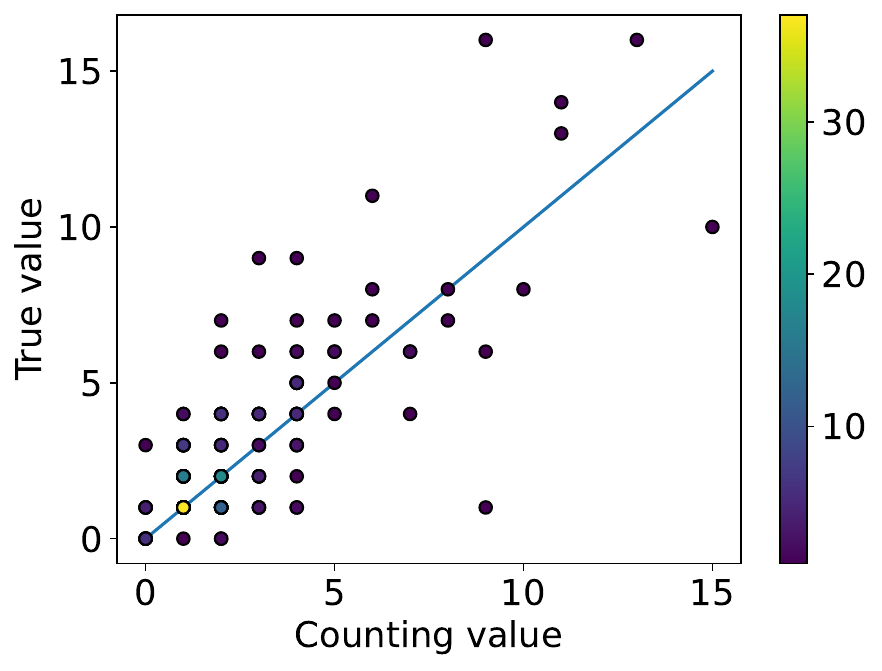}
        \caption{\intern 78B}
        \label{fig:internvl_stmf_count}
    \end{subfigure}
    \begin{subfigure}{.23\textwidth}
        \centering
        \includegraphics[width=0.8\linewidth]{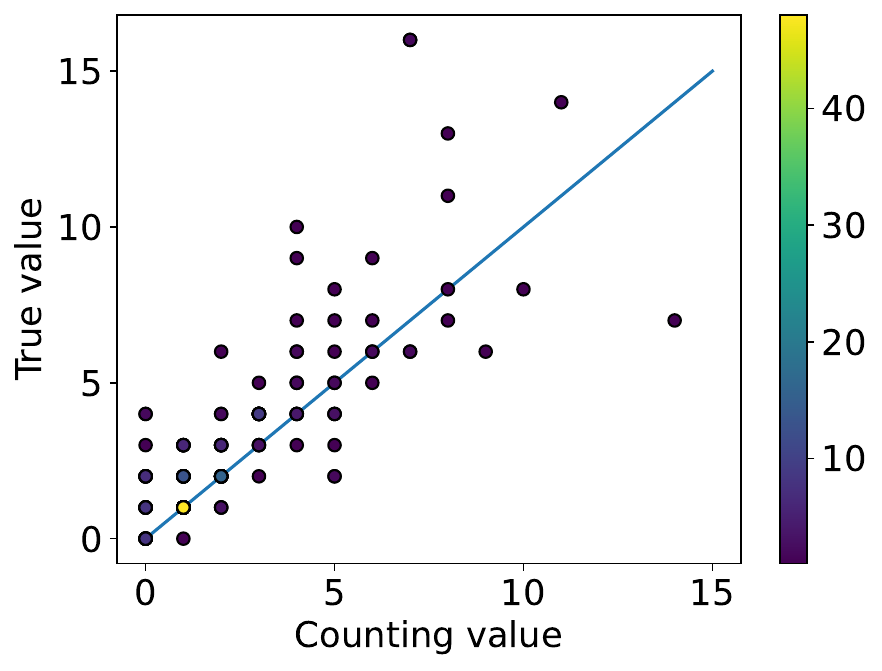}
        \caption{\oAI{} \othree}
        \label{fig:o3_stmf_count}
    \end{subfigure}
    \caption{Scatter plots of true versus counts estimated by \acp{vlm}. Blue lines indicate correct estimation. To address overlapping points with identical true and estimated values, color gradients represent point density for improved visual clarity.}
    \Description{A detailed description of the image content.}
    \label{fig:models_stmf_count}
\end{figure}

\begin{figure}[ht]
    \centering
    \includegraphics[width=0.8\linewidth]{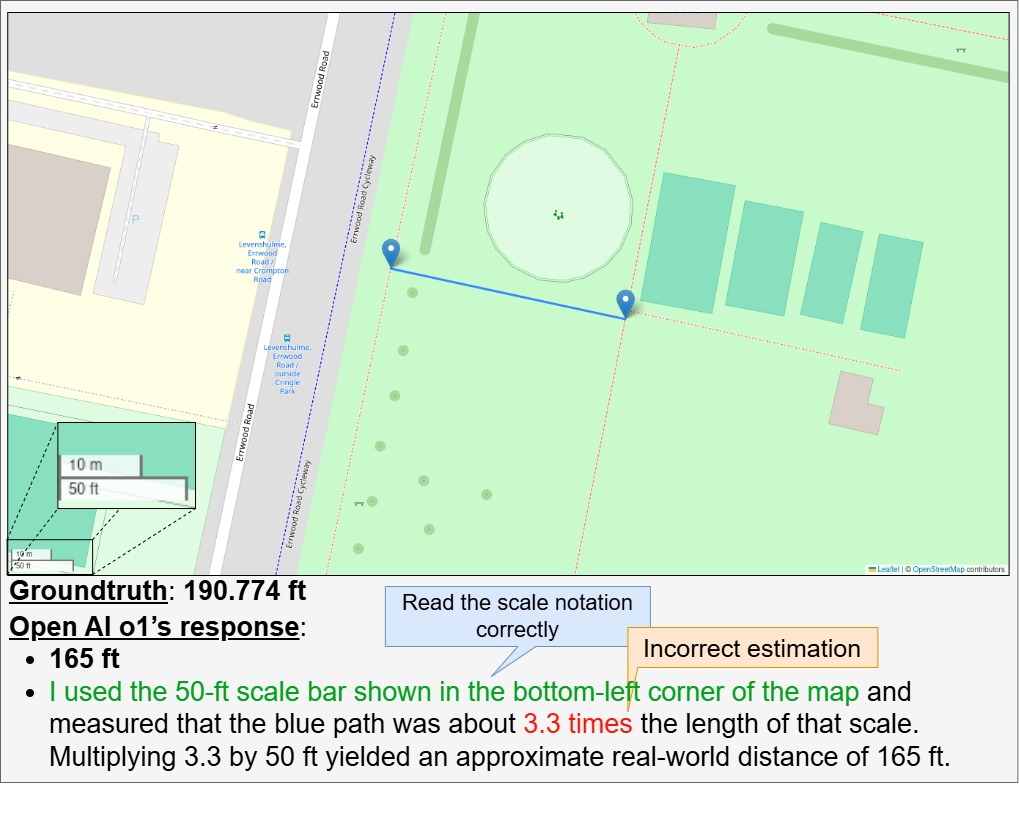}
    \caption{Example of \OAI{} \Oone's response in the \acf{rlest} \subtask, demonstrating the ability of reading the scale notation but failing to correctly measure the path's length based on the scale unit.}
    \Description{A detailed description of the image content.}
    \label{fig:example_merit}
\end{figure}

\begin{figure}[ht]
    \begin{subfigure}{.23\textwidth}
        \centering
        \includegraphics[width=1.\linewidth]{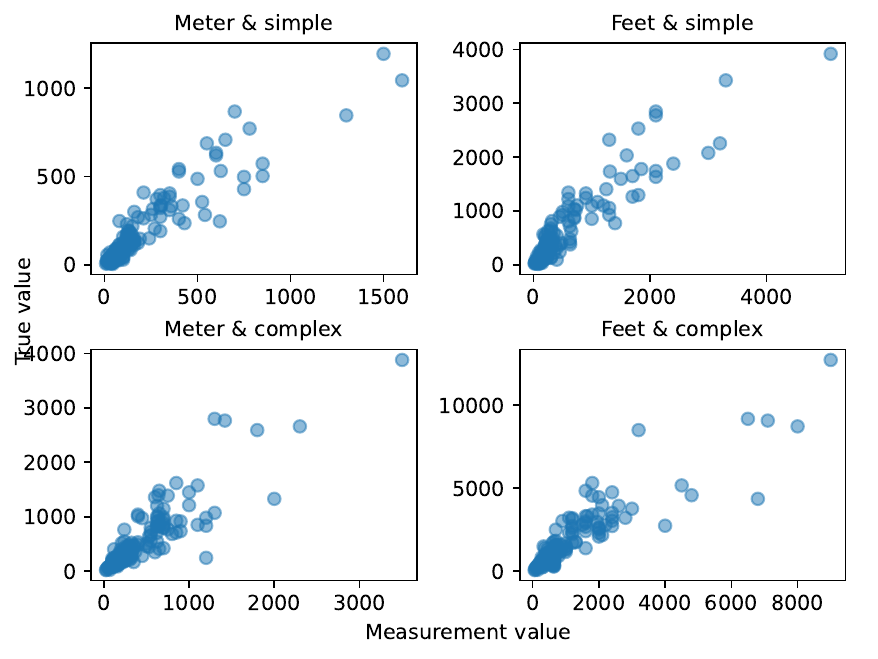}
        \caption{\oAI{} \oone}
        \label{fig:o1_lenmes}
    \end{subfigure}
    \begin{subfigure}{.23\textwidth}
        \centering
        \includegraphics[width=1.0\linewidth]{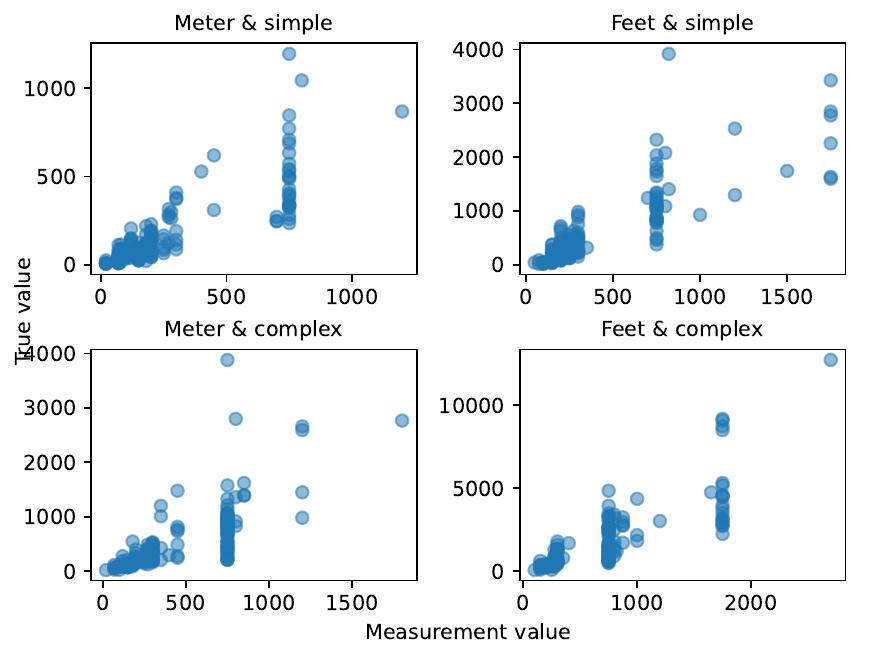}
        \caption{\qwen 72B}
        \label{fig:qwen_lenmes}
    \end{subfigure}
    \caption{Scatter plots comparing model’s proposed measurements with true route lengths in the \ac{rlest} \subtask.}
    \Description{A detailed description of the image content.}
    \label{fig:models_scatter_lenmes}
\end{figure}

\begin{figure}[ht]
    \centering
    \includegraphics[width=1.0\linewidth]{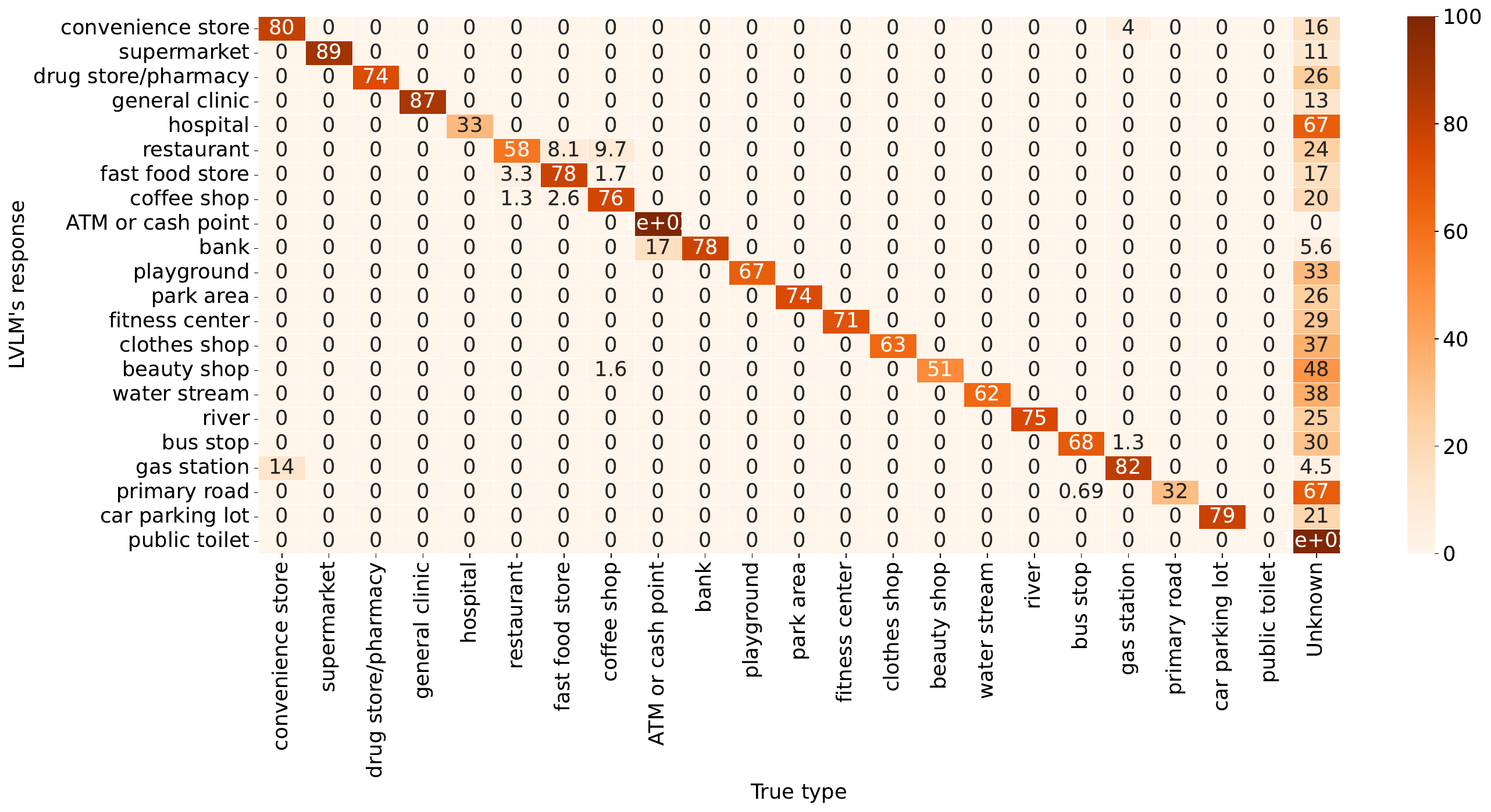}
    \caption{Confusion matrix of \oAI{} \oone’s responses to the \texttt{name-listing} \question in the \acf{mtmf} \subtask. Each cell represents a percentage, where the values of each row sum to $100$\%. The final column, ``Unknown'', corresponds to names that do match any of the considered \ac{mapfeat} types, and may potentially result from \ac{ocr} errors or hallucinations.}
    \Description{A detailed description of the image content.}
    \label{fig:confus_o1}
\end{figure}

\begin{figure}[ht]
    \centering
    \includegraphics[width=0.85\linewidth]{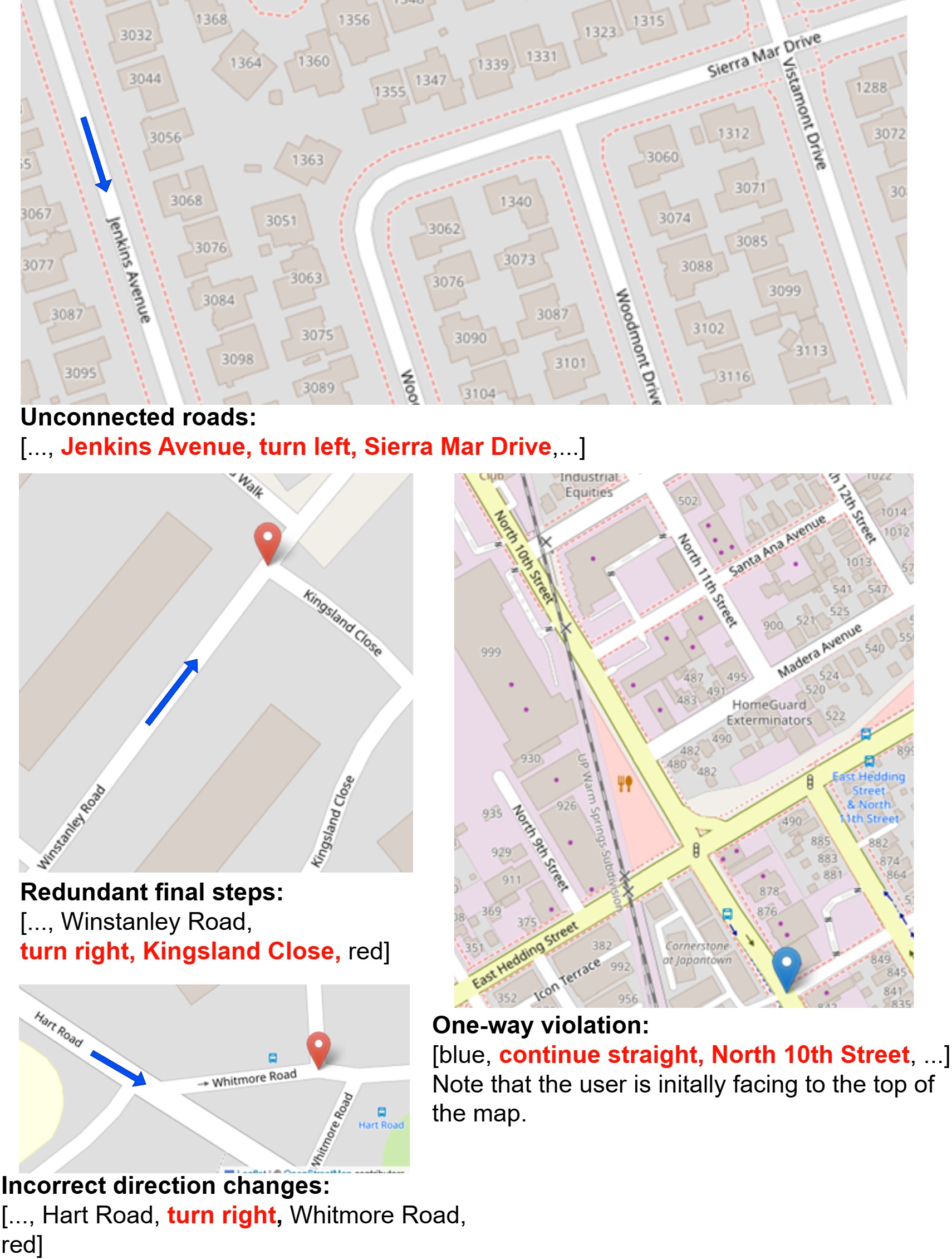}
    \caption{Examples of issues found in \Othree's responses in the \acf{srnav} \subtask. Notes: due to space constraints, the \aclp{cm} are cropped from the original inputs. And, blue arrows are overlaid to indicate the model's proposed route.}
    \Description{A detailed description of the image content.}
    \label{fig:example_navi}
\end{figure}

\begin{figure}[ht]
    \centering
    \includegraphics[width=1.\linewidth]{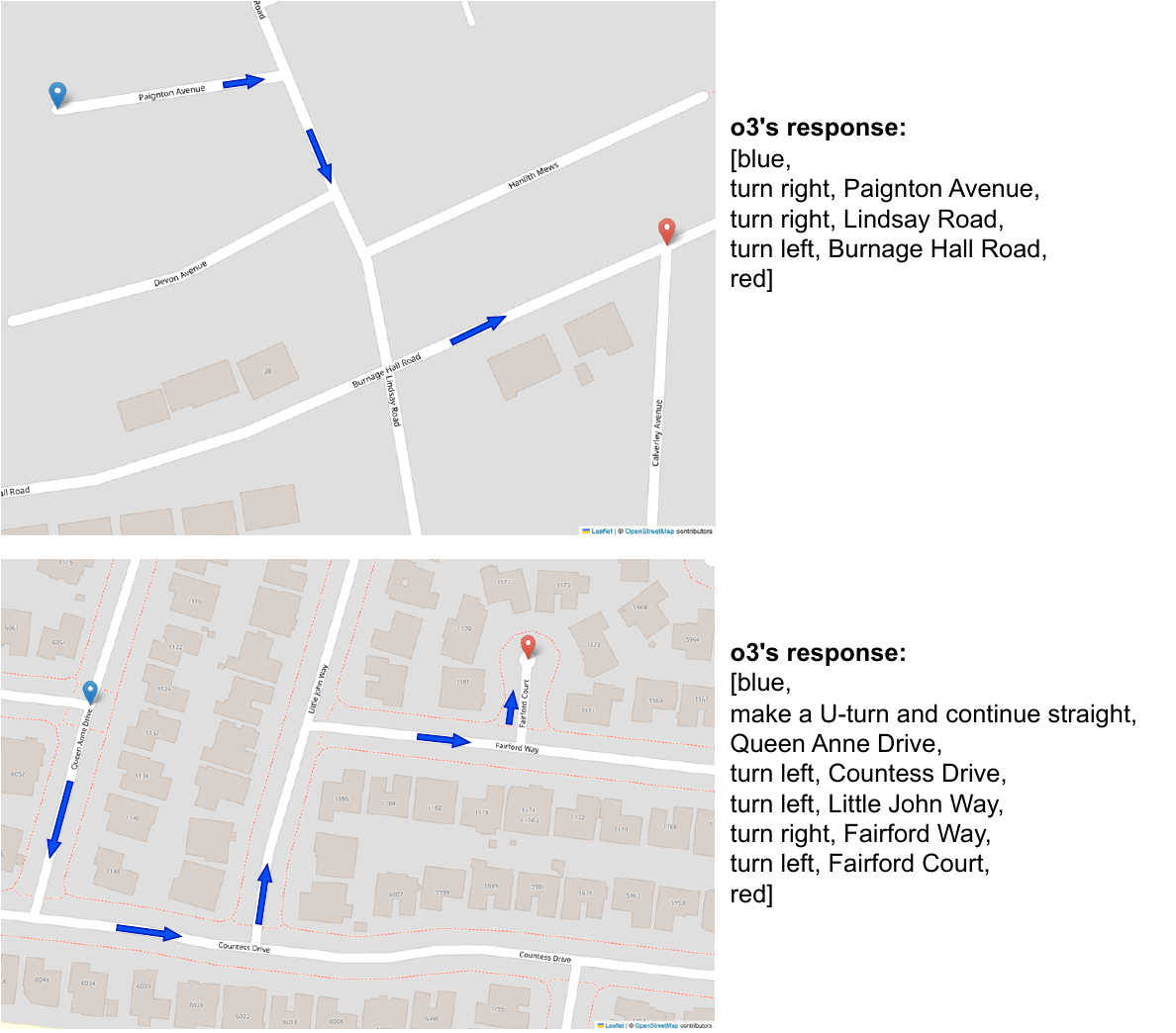}
    \caption{Examples of successful shortest routes generated by \Othree in the \acf{srnav} \subtask. Notes: due to space constraints, the \aclp{cm} are cropped from the original inputs. And, blue arrows are overlaid to indicate the model's proposed route.}
    \Description{A detailed description of the image content.}
    \label{fig:example_navi_success}
\end{figure}

The list of \acp{mapfeat} is provided in Table~\ref{tab:select_map_features}.
Figure~\ref{fig:lenme_distri} and~\ref{fig:step_distribution} show the distribution plots described in Section 3.1 for MTMF and SRN, respectively.
Figure~\ref{fig:models_stmf_count} illustrates  the models’ tendency to undercount.
Table~\ref{tab:compare_stmf_mtmf} further compares STMF and MTMF results, as discussed in Section 4.3, while Table~\ref{tab:task2_cot} reports results with and without CoT.
Figure~\ref{fig:confus_o1} presents a confusion matrix highlighting potential misrecognition or OCR errors.
Figure~\ref{fig:models_scatter_lenmes} compares real and estimated lengths across models as mentioned in Section 4.3.
Figure~\ref{fig:example_merit} presents an example of correctly interpreting the scale bar.
Finally, Figure~\ref{fig:example_navi} and~\ref{fig:example_navi_success} illustrate representative failure cases and correctly generated routes, respectively.

\begin{table}[t]
    \centering
    \caption{\small{Description of map features utilized in the \acf{stmf} and the \acf{mtmf} \subtasks.}}
    \resizebox{0.65\columnwidth}{!}{%
    \begin{tabular}{llccc}
    \hline
         \textbf{No.}& \textbf{Type} & \textbf{Category}  &Class&\textbf{Image}\\ \hline
         1& Convenience store & Symbol &Shop 
&\includegraphics[width=0.35cm]{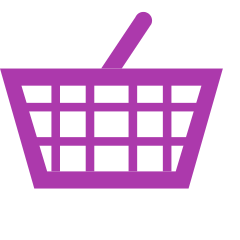}\\ \hline
         2& Supermarket & Symbol &Shop 
&\includegraphics[width=0.35cm]{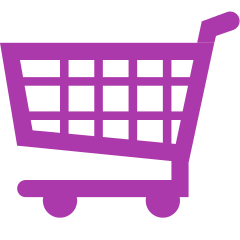}\\ \hline 
         3& Drug store/pharmacy & Symbol &Amenity 
&\includegraphics[width=0.35cm]{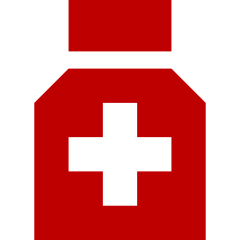}\\ \hline  
 4&General clinic & Symbol &Amenity 
&\includegraphics[width=0.35cm]{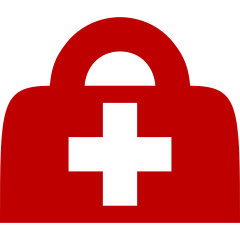}\\ \hline 
 5&Hospital & Symbol &Amenity 
&\includegraphics[width=0.35cm]{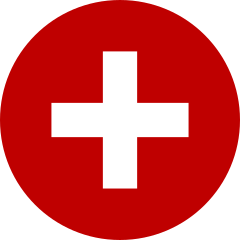}\\ \hline
 6&Restaurant & Symbol &Amenity 
& \includegraphics[width=0.35cm]{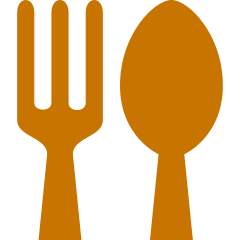}\\ \hline
 7&Fast food store & Symbol &Amenity 
& \includegraphics[width=0.35cm]{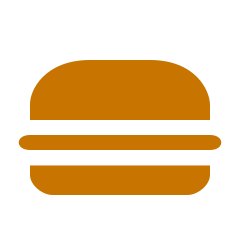}\\ \hline
 8&Coffee shop & Symbol &Amenity 
&\includegraphics[width=0.35cm]{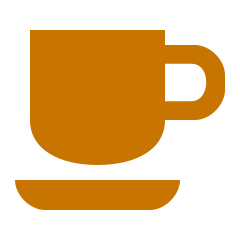}\\ \hline
 9&ATM or cash point & Symbol &Amenity 
&\includegraphics[width=0.35cm]{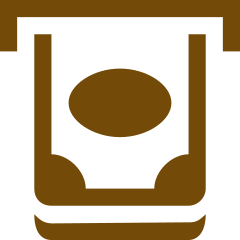}\\ \hline
 10&Bank & Symbol &Amenity 
&\includegraphics[width=0.35cm]{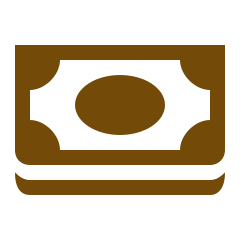}\\ \hline
 11&Playground & Symbol &Leisure 
&\includegraphics[width=0.35cm]{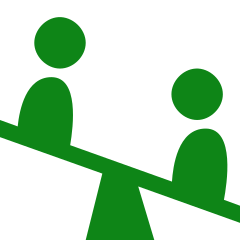}\\ \hline
 12&Park area & Area &Leisure 
&\includegraphics[width=0.35cm]{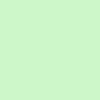}\\ \hline
 13&Fitness center & Symbol &Leisure 
&\includegraphics[width=0.35cm]{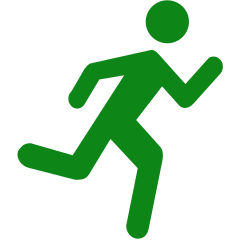}\\ \hline
 14&Clothes shop & Symbol &Shop 
&\includegraphics[width=0.35cm]{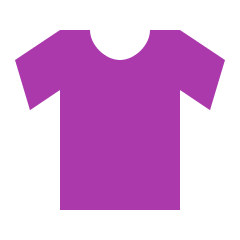}\\ \hline
 15&Beauty shop & Symbol &Shop 
&\includegraphics[width=0.35cm]{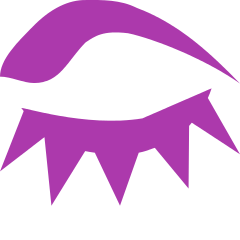}\\ \hline
 16&Water stream & Line &Waterway&\includegraphics[width=0.4cm]{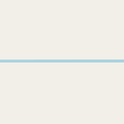}\\ \hline
 17&River & Line &Waterway&\includegraphics[width=0.4cm]{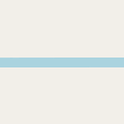}\\ \hline
 18&Bus stop & Symbol &Highway 
&\includegraphics[width=0.35cm]{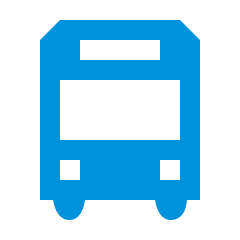}\\ \hline
 19&Gas station & Symbol &Amenity 
&\includegraphics[width=0.35cm]{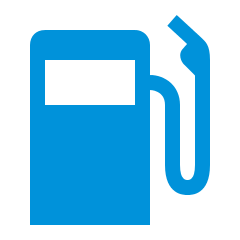}\\ \hline
 20&Primary road & Line &Highway 
&\includegraphics[width=0.4cm]{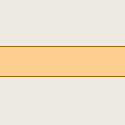}\\ \hline
 21&Car parking lot & Symbol &Amenity 
&\includegraphics[width=0.30cm]{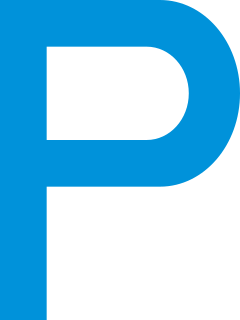}\\ \hline
 22&Public toilet & Symbol &Amenity &\includegraphics[width=0.35cm]{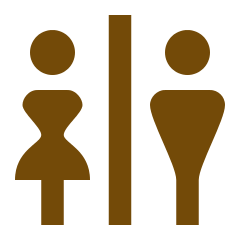}\\ \hline
    \end{tabular}
    }
    
    \label{tab:select_map_features}
\end{table}

\begin{table}[!h]
    \centering
    \caption{Comparison in performance between \acf{stmf} and \acf{mtmf} tasks, measured on \ac{stmf}.}
    \resizebox{0.8\columnwidth}{!}{%
 \begin{tabular}{l|r|r|r|r}
 \hline
 & \multicolumn{2}{c|}{STMF}& \multicolumn{2}{c}{MTMF}\\
        \hline
         \textbf{Model} &\textbf{Counting}&\textbf{Name listing} & \textbf{Counting}&\textbf{Name listing} 
\\
           & a$r^2$ ($\uparrow$)& amF1 ($\uparrow$) & a$r^2$ ($\uparrow$)&amF1 ($\uparrow$) \\ \hline
          Llama 3.2 11B
& \textbf{0.143}& \textbf{0.342}& -0.737&0.163
\\
 InternVL2.5 8B& \textbf{0.095}& \textbf{0.407}& -0.112&0.285
\\
 LlaVa-OV 7B& \textbf{0.528}
& \textbf{0.340}& -0.167&0.109
\\
 QwenVL 2.5 7B
& \textbf{0.342}& \textbf{0.624}
& 0.042&0.482
\\ \hline
 Llama 4 Scout
& 0.266& \textbf{0.663}& \textbf{0.270}&0.524
\\
 Llama 3.2 90B
& \textbf{0.312}& \textbf{0.542}& -0.216&0.356
\\
 InternVL2.5 78B
& \textbf{0.618}
& \textbf{0.519}& 0.289&0.455
\\
 LlaVa-OV 72B&  \textbf{0.490}& \textbf{0.537}& -0.274&0.311
\\
 QwenVL 2.5  72B
& \textbf{0.401}& \textbf{0.706}
& 0.311&0.610
\\ \hline
 GPT-4(V)
& \textbf{0.359}&  \textbf{0.687}& -0.166&0.491\\
 GPT-4o
& \textbf{0.626}& \textbf{0.821}& 0.142&0.592\\
 o1
& \textbf{0.581}& \textbf{0.813}& 0.486&0.791
\\ 
          o3
& \textbf{0.592}&  \textbf{0.847}& 0.474&0.778
\\
 Gemini 2.5 Pro
& \textbf{0.761}
& \textbf{0.894}
& 0.584&0.808
\\
          Claude 3.7 Sonnet
& \textbf{0.624}& \textbf{0.714}& 0.520&0.633
\\ \hline
    \end{tabular}
    }
\label{tab:compare_stmf_mtmf}
\end{table}

\begin{table}[ht]
    \centering
    \caption{\small{Comparison on the performance with (\textit{w/}) and without (\textit{w/o}) the \acf{cot} strategy. Experiments were performed with two advanced proprietary models that lack native extended thinking ability.}}
    \resizebox{\columnwidth}{!}{%
    \begin{tabular}{l|r|r|r|r|r|r|r|r}
        \hline
         & \multicolumn{4}{c}{\textbf{Simple}}& \multicolumn{4}{c}{\textbf{Complex}}\\ 
         & \multicolumn{2}{c}{\textbf{Meter}}& \multicolumn{2}{c}{\textbf{Feet}}& \multicolumn{2}{c}{\textbf{Meter}}& \multicolumn{2}{c}{\textbf{Feet}}\\ 
         & RMSE ($\downarrow$)& $r^2$ ($\uparrow$)& RMSE ($\downarrow$)& $r^2$ ($\uparrow$)& RMSE ($\downarrow$)& $r^2$ ($\uparrow$)& RMSE ($\downarrow$)& $r^2$ ($\uparrow$) \\ \hline
         \gptV{} w/o  CoT & \textbf{146.79} & \textbf{0.53} & 638.93 & 0.17 & 468.02 & 0.37 & 2040.8 & -0.11 \\
         \gptV{} w/  CoT & 271.15 & -0.61 & \textbf{549.96} & \textbf{0.38} & \textbf{449.77} & \textbf{0.42} & \textbf{1678.36} & \textbf{0.24}\\ \hline
         \gptO{} w/o CoT & 249.73& -0.37& 966.10 & -0.9 & 534.51 & 0.17 & 1733.21 & 0.19 \\
         \gptO{} w/  CoT & \textbf{208.86} & \textbf{0.04} & \textbf{526.32} & \textbf{0.44} & \textbf{314.93} & \textbf{0.71} & \textbf{1344.83} & \textbf{0.51}\\\hline
    \end{tabular}
    }
\label{tab:task2_cot}
\end{table}

\end{document}